\documentclass{article}

\usepackage[final]{corl_2021}

\usepackage[numbers]{natbib}
\usepackage{multicol}

\usepackage{breakurl}

\usepackage{parskip}

\usepackage{dsfont}
\usepackage{amsfonts}
\usepackage{amsmath}
\usepackage{amssymb}
\usepackage{bbm}
\usepackage[ruled,noend,linesnumbered]{algorithm2e}
\usepackage{setspace}
\usepackage{tikz}
\usepackage{booktabs}
\usepackage{multirow}
\usepackage{graphicx}
\usepackage{placeins}

\usepackage{enumitem}

\let\vec\mathbf

\renewcommand{\d}{\ensuremath{\>\mathrm{d}}}

\title{\textsc{RoCUS}: Robot Controller\\Understanding via Sampling}

\author{
\begin{tabular}{cccc}
    Yilun Zhou & Serena Booth & Nadia Figueroa & Julie Shah\\
\end{tabular}\vspace{0.03in}\\
MIT CSAIL\vspace{0.03in}\\
\texttt{\{yilun, serenabooth, nadiafig, julie\_a\_shah\}@csail.mit.edu}
}

\usepackage{wrapfig}
\usepackage{appendix}

\newcommand{\modelname}{\textsc{RoCUS}}

\begin{document}
\maketitle

\vspace{-0.3in}
{\centering \url{https://yilunzhou.github.io/RoCUS}\par}

\begin{abstract}
As robots are deployed in complex situations, engineers and end users must develop a holistic understanding of their behaviors, capabilities, and limitations. Some behaviors are directly optimized by the objective function. They often include success rate, completion time or energy consumption. Other behaviors---e.g., collision avoidance, trajectory smoothness or motion legibility---are typically emergent but equally important for safe and trustworthy deployment. Designing an objective which optimizes every aspect of robot behavior is hard. In this paper, we advocate for systematic analysis of a wide array of behaviors for holistic understanding of robot controllers and, to this end, propose a framework, \modelname{}, which uses Bayesian posterior sampling to find situations where the robot controller exhibits user-specified behaviors, such as highly jerky motions. We use \modelname{} to analyze three controller classes (deep learning models, rapidly exploring random trees and dynamical system formulations) on two domains (2D navigation and a 7 degree-of-freedom arm reaching), and uncover insights to further our understanding of these controllers and ultimately improve their designs. 

\end{abstract}


\section{Introduction}
\label{sec:intro}

In 2018, after a confluence of failures, an autonomous vehicle (AV) struck and killed a pedestrian for the first time. In the run-up to this fateful event, the responsible company had reportedly been trying to improve the AV ``ride experience" by emphasizing non-critical behaviors---such as the smoothness of the ride~\cite{ubercrash}. This event reflects the long-standing challenge in robotics: designing an appropriate objective which considers both safety-critical and non-critical behaviors. 
When crafting an objective, it is virtually impossible to proactively account for all potential controller behaviors, and some priorities may even be in conflict with one another~\cite{rasouli2019autonomous}. In practice, any given robot behaviors may be specified, unspecified, or even misspecified~\cite{bobu2020quantifying}, so extensive testing and evaluation is a critical component of designing and assessing robot controllers---especially those using black-box models such as deep neural networks.

A common testing procedure focuses on finding extreme and edge cases of controller failure. For example, a tester might use this procedure to find that the AV swerves very badly when encountering a farm animal while traveling at 60mph. Finding such extreme and edge cases is well-studied within both traditional software testing paradigms~\cite{myers2004art} and more recent adversarial perturbation testing methods~\cite{goodfellow2014explaining}. However, we argue that an equally, if not more, important form of testing should focus on \emph{representative} scenarios, which considers the likelihood of encountering these scenarios. For example, if this AV is going to be deployed exclusively in New York City, the above example is largely unhelpful: cars rarely travel at 60mph in the city, and are very unlikely to encounter farm animals. Instead, the tester may prefer to know that the car swerves---though not as substantively---at lower speeds when a pedestrian steps toward it. Finding representative scenarios is often overlooked, but is especially useful for robotics. This is the focus of this paper.

Explicit mathematical analysis of robot controllers is implausible given the high dimensionality of the configuration space and the potential black-box representation of a learned controller. With access to an environment simulator, though, a straightforward testing approach is to roll out the robotic controller on various environments (e.g. road conditions under different weather and congestion, with or without farm animals or pedestrians, etc.), and analyze those rollouts that exhibit a specified behavior---like excessive swerving. However, with too few environments, we risk missing the condition(s) that triggers the target behavior most saliently. With too many environments, all the most salient rollouts would be close to the global maximum at the expense of diversity and coverage. For example, if a farm animal causes the most swerving, followed by a pedestrian and a dangling tree branch, using too few environments may only find the pedestrian and the tree branch while using too many would result in an exclusive focus on the farm animal. Neither case helps the human develop a correct mental model of the AV's behavior.

\begin{wrapfigure}{rt}{0.5\textwidth}
    \centering
    \includegraphics[height=85px,trim={100px 37px 90px 40px},clip]{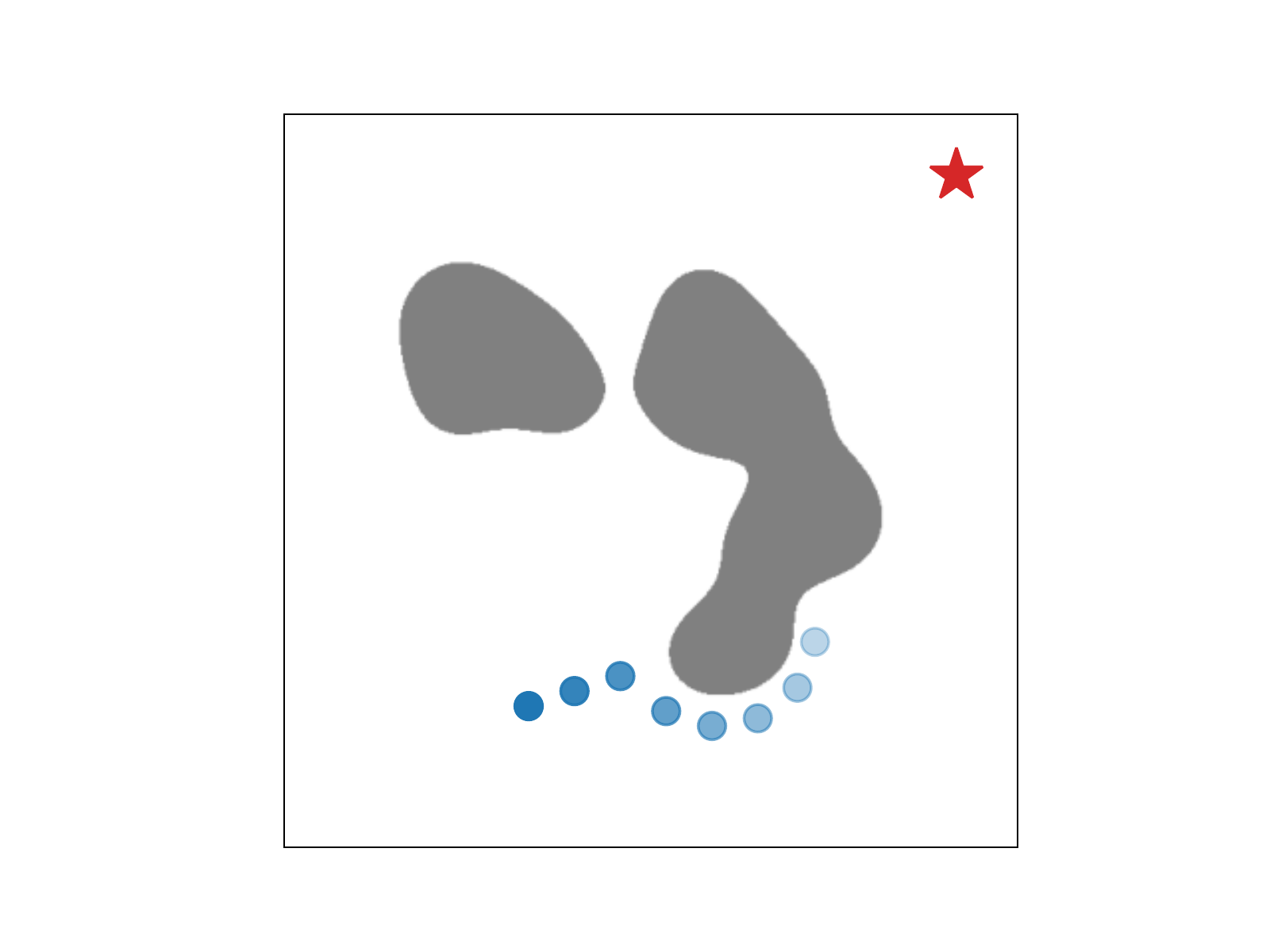}
    \includegraphics[height=85px, trim=0.5in 0 0.5in 0, clip]{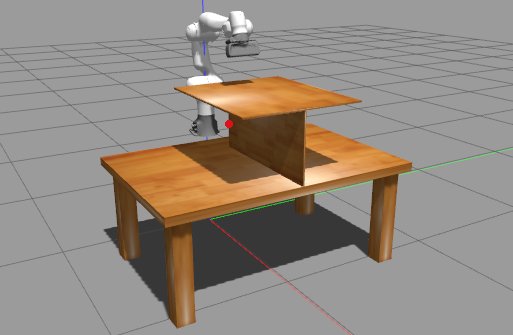}
    \caption{Two use case demos of \modelname{}: 2D navigation (left) and 7DoF arm reaching (right). }
    \label{fig:task-overview}
\end{wrapfigure}

To address this, we introduce Robot Controller Understanding via Sampling (\modelname{}), a method to enable systematic behavior inspection. \modelname{} finds scenarios that are both inherently likely and elicit specified behaviors by formulating the problem as one of Bayesian posterior inference. Analyzing these scenarios and the resulting trajectories can help developers better understand the robot behaviors, and allow them to iterate on algorithm development if undesirable ones are revealed.

We use \modelname{} to analyze three controllers on two common robotics tasks (Fig.~\ref{fig:task-overview}). For a 2D navigation problem, we consider imitation learning (IL) \cite{argall2009survey}, dynamical system (DS) \cite{huber2019avoidance}, and rapidly-exploring random tree (RRT) \cite{lavalle2006planning}. For a 7DoF arm reaching problem, we consider reinforcement learning (RL) \cite{sutton1998introduction}, as well as the same DS and RRT controllers. For each problem and controller, we specify several behaviors and visualize representative scenarios and trajectories that elicit those behaviors. Through this analysis, we uncover insights that would be hard to derive analytically and thus complement our mathematical understanding of the controllers. Moreover, we include a case study on how to improve a controller based on new insights from \modelname{}. As such, \modelname{} is a step towards the broader goal of building more accurate human mental models and enabling holistic evaluation of robot behaviors.

\section{Related Work}
Our work lies at the intersection of efforts to understand complex model behaviors and those to benchmark robot performance. Methods to understand, interpret, and explain model behaviors are now commonplace in the machine learning community. \citet{mitchell2019model} introduced Model Cards, a model analysis mechanism which breaks down model performance for data subsets. In natural language processing,~\citet{ribeiro2020beyond} introduced a checklist for holistic evaluation of model capabilities and test case generation. \citet{booth2020bayes} introduced \textsc{Bayes-TrEx}, a Bayesian inference framework for sampling specified classifier behaviors.  In robotics, \citet{fan2020parameter} introduced a verification framework for assessing machine behavior by sampling parameter spaces to find temporal logic-satisfying behaviors. Other efforts aim to summarize robot policies, trading off factors like brevity, diversity and completeness~\cite{hayes2017improving, lage2019exploring}. All of these works have a shared underlying theme: treating the black box as immutable and performing downstream analyses of machine behavior~\cite{rahwan2019machine}. \modelname{} shares this theme and, similar to \textsc{Bayes-TrEx} \citep{booth2020bayes}, searches for instances which exhibit target behaviors to inform accurate human mental models. 

While the need for benchmarking robot performance is often expressed~\cite{mahler2018guest,murali2019pyrobot,james2020rlbench}, these efforts usually operate on distributions of trajectories or randomly selected trajectories, and the accompanying metrics are typically task-completion based without consideration of implicit performance factors. \citet{anderson2018evaluation} put forth a recommendation of using \emph{success weighted by path length} for navigation tasks---a task-completion metric. \citet{cohen2012generic} and \citet{moll2015benchmarking} introduced suites of metrics for comparing motion planning approaches, and \citet{lagriffoul2018platform} presented a set of task and motion planning scenarios and metrics. Again, all of these proposed metrics are based solely on task completion. \citet{lemme2015open} proposed a set of performance measures for reaching tasks, which are either task-completion based or require a costly human motion ground truth. Our contribution is distinct in two ways. First, we propose to sample specific trajectories which communicate controller behaviors instead of reporting metrics averaged over distributions of trajectories. Second, we introduce metrics which draw on these prior works while also including essential alternative and typically emergent quality factors, like motion jerkiness and legibility~\cite{dragan2013legibility}.

\section{\modelname{}}
\label{sec:method}

\begin{wrapfigure}{r}{0.4\textwidth}
    \centering
    \vspace{-0.3in}
    \tikzset{every picture/.style={line width=0.75pt}} 

\begin{tikzpicture}[x=0.75pt,y=0.75pt,yscale=-1,xscale=1]

\draw   (156.31,68.59) .. controls (156.31,62.19) and (161.5,57) .. (167.91,57) .. controls (174.31,57) and (179.5,62.19) .. (179.5,68.59) .. controls (179.5,75) and (174.31,80.19) .. (167.91,80.19) .. controls (161.5,80.19) and (156.31,75) .. (156.31,68.59) -- cycle ;
\draw   (215.31,68.59) .. controls (215.31,62.19) and (220.5,57) .. (226.91,57) .. controls (233.31,57) and (238.5,62.19) .. (238.5,68.59) .. controls (238.5,75) and (233.31,80.19) .. (226.91,80.19) .. controls (220.5,80.19) and (215.31,75) .. (215.31,68.59) -- cycle ;

\draw   (275.31,68.59) .. controls (275.31,61.19) and (280.5,56) .. (286.91,56) .. controls (293.31,56) and (298.5,61.19) .. (298.5,68.59) .. controls (298.5,74) and (293.31,79.19) .. (286.91,79.19) .. controls (280.5,79.19) and (275.31,74) .. (275.31,68.59) -- cycle ;

\draw   (335.31,68.59) .. controls (335.31,61.19) and (340.5,56) .. (346.91,56) .. controls (353.31,56) and (358.5,61.19) .. (358.5,68.59) .. controls (358.5,74) and (353.31,79.19) .. (346.91,79.19) .. controls (340.5,79.19) and (335.31,74) .. (335.31,68.59) -- cycle ;
\draw    (179.5,68.59) -- (212.31,68.59) ;
\draw [shift={(215.31,68.59)}, rotate = 180] [fill={rgb, 255:red, 0; green, 0; blue, 0 }  ][line width=0.08]  [draw opacity=0] (10.72,-5.15) -- (0,0) -- (10.72,5.15) -- (7.12,0) -- cycle    ;
\draw    (238.5,68.59) -- (272.31,68.59) ;
\draw [shift={(275.31,68.59)}, rotate = 538.44] [fill={rgb, 255:red, 0; green, 0; blue, 0 }  ][line width=0.08]  [draw opacity=0] (10.72,-5.15) -- (0,0) -- (10.72,5.15) -- (7.12,0) -- cycle    ;
\draw    (298.5,68.59) -- (332.31,68.59) ;
\draw [shift={(335.31,68.59)}, rotate = 180] [fill={rgb, 255:red, 0; green, 0; blue, 0 }  ][line width=0.08]  [draw opacity=0] (10.72,-5.15) -- (0,0) -- (10.72,5.15) -- (7.12,0) -- cycle    ;
\draw    (167.91,57) .. controls (207.31,27.45) and (242.43,27.67) .. (284.96,54.74) ;
\draw [shift={(286.91,56)}, rotate = 213.12] [fill={rgb, 255:red, 0; green, 0; blue, 0 }  ][line width=0.08]  [draw opacity=0] (10.72,-5.15) -- (0,0) -- (10.72,5.15) -- (7.12,0) -- cycle    ;
\draw  [dash pattern={on 4.5pt off 4.5pt}] (305.5,55.59) .. controls (305.5,51.18) and (309.08,47.59) .. (313.5,47.59) -- (362.31,47.59) .. controls (366.73,47.59) and (370.31,51.18) .. (370.31,55.59) -- (370.31,79.59) .. controls (370.31,84.01) and (366.73,87.59) .. (362.31,87.59) -- (313.5,87.59) .. controls (309.08,87.59) and (305.5,84.01) .. (305.5,79.59) -- cycle ;

\draw (167,68.59) node   [align=center] {$\displaystyle t$};
\draw (227,68.59) node [inner sep=0.75pt]   [align=center] {$\displaystyle \tau $};
\draw (287,68) node [inner sep=0.75pt]   [align=center] {$\displaystyle b$};
\draw (347,67) node [inner sep=0.75pt]  [align=center] {$\displaystyle \hat{b}$};
\end{tikzpicture}
    \vspace{-6mm}
    \caption{The graphical model for the inference problem of finding tasks $t$ and trajectories $\tau$ which exhibit specific behaviors $b$. The dashed box indicates the relaxed formulation (Eq. \ref{eq:bhat-and-relaxed-posterior}).}
    \label{fig:inference-diagram}
\end{wrapfigure}
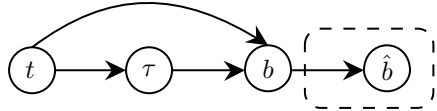

At a high level, \modelname{} helps users understand robotic controllers via representative scenarios that exhibit various specified behaviors. It solves this by directly incorporating the distribution of scenarios, formally called \textit{tasks}, into a Bayesian inference framework as shown in Fig.~\ref{fig:inference-diagram}. A robotic problem is represented by a distribution $\pi(t)$ of individual tasks $t$. For example, a navigation problem may have $\pi(t)$ representing the distribution over target locations and obstacle configurations. Given a specific task $t$, the controller under study induces a distribution $p(\tau | t)$ of possible trajectories $\tau$. If both the controller and the transition dynamics are deterministic, $p(\tau | t)$ reduces to a $\delta$-function at the induced trajectory $\tau$. Stochasticity in either the controller (e.g., RRT) or the dynamics (e.g., uncertain outcome from an action) can result in $\tau$ being random. Finally, a behavior function $b(\tau, t)$ computes the behavior value of the trajectory---for example, the motion jerkiness. Some behaviors only depend on the trajectory and not the task, but we use $b(\tau, t)$ for consistency. Sec.~\ref{sec:metrics} presents a list of behaviors.

The discussion on behavior in Sec.~\ref{sec:intro} is informal and implicitly combines two related but different concepts. The first concept is the behavior function $b(\tau, t)$ discussed above. The second is the specified target: for the swerving example, we are particularly interested in \textit{maximal} behavior values. Thus, the target value can be thought of as $+\infty$. This inference problem uses the \textit{maximal} mode of \modelname{}. 
In other cases, we are also interested in tasks and trajectories whose behaviors \textit{matches} a target. For example, we want to find road conditions that lead to a daily commute time of an hour, where the behavior is the travel time. This inference problem uses the \emph{matching} mode. 
Since matching mode is conceptually simpler, we present it first, followed by maximal mode. The sampling procedure is the same for both modes and presented last in Alg.~\ref{alg:mh}. 

\subsection{Matching Mode}
The exact objective is to find tasks and trajectories that exhibit user-specified behaviors $b^*$: 
\begin{align}
t, \tau \sim p(t, \tau | b = b^*)\propto p(b = b^* | t, \tau)\pi(\tau | t)\pi(t). 
\end{align}
In most cases this posterior does not admit direct sampling, and an envelope distribution is not available for rejection sampling. Markov-Chain Monte-Carlo (MCMC) sampling does not work either: since the posterior is only non-zero on a very small or even measure-zero set, a Metropolis-Hastings (MH) sampler \cite{hastings1970monte} can get stuck in the zero-density region. 
Similar to the \textsc{Bayes-TrEx} formulation~\cite{booth2020bayes}, we relax it using a normal distribution formulation as shown in Fig.~\ref{fig:inference-diagram}:
\begin{align}
\widehat b | b \sim \mathcal{N}(b, \sigma^2) \,\,\quad
t, \tau \sim p(t, \tau | \widehat b=b^*)\propto p(\widehat b=b^* | t, \tau)p(\tau | t) \pi(t). \label{eq:bhat-and-relaxed-posterior}
\end{align}
This relaxed posterior is non-zero everywhere $\pi(t)$ is non-zero and provides useful guidance to an MH sampler. While $\sigma$ is a hyper-parameter in \textsc{Bayes-TrEx}~\cite{booth2020bayes}, we instead choose $\sigma$ such that
\begin{align}
\int_{b^*-\sqrt{3}\sigma}^{b^*+\sqrt{3}\sigma} p(b)\d b = \alpha, \,\,\,\mathrm{with} \,\,\, p(b) = \int_t\int_{\tau} p(\tau|t)\pi(t)\mathbbm{1}_{b(\tau, t)=b}\d {\tau}\d t  \label{eq:alpha-and-b-marginal}
\end{align}
being the marginal distribution of $b(\tau, t)$, which can be estimated by trajectory roll-outs. This formulation has two desirable properties. First, it is scale-invariant with respect to $b(\tau, t)$, e.g. measured under different units like meters vs. centimeters. Second, the hyper-parameter $\alpha\in [0, 1]$ has the intuitive interpretation of the approximate ``volume'' of posterior samples $t, \tau \mid \widehat b=b^*$ under the marginal $p(t, \tau) = p(\tau | t)\pi(t)$, a notion of their representativeness. Details are derived in App. \ref{supp:scale-invariance}.

\subsection{Maximal Mode}
In this mode, \modelname{} finds trajectories that lead to maximal behavior values: $b^*\rightarrow \pm\infty$. It can also be used for finding minimal behavior values by negating the behavior. The posterior formulation is: 
\begin{alignat}{2}
    b_0 = \frac{b-\mathbb E[b]}{\sqrt{\mathbb{V}[b]}},
    \quad \beta = \frac{1}{1 + e^{-b_0}}, \quad
    \widehat\beta \sim \mathcal{N}\left(\beta, \sigma^2\right), 
    \quad t, \tau  \sim p(t, \tau | \widehat\beta=1),\label{eq:b-standardize-sigmoid-and-maximal-posterior}
\end{alignat}
where $\mathbb E[b]$ and $\mathbb{V}[b]$ are the mean and variance of the marginal $p(b)$. $\sigma$ is chosen such that
\begin{align}
    \int_{1-\sqrt{3}\sigma}^1 p(\beta)\d\beta=\alpha, \label{eq:alpha-and-b-marginal-beta}
\end{align}
where $p(\beta)$ is the marginal distribution similar to Eq.~\ref{eq:alpha-and-b-marginal}. If $p(b)$ is normal, $p(\beta)$ is logit-normal. This formulation is again scale-invariant and has the same ``volume'' interpretation for $\alpha$ (App.~\ref{supp:scale-invariance}). 

\subsection{Posterior Sampling}

\begin{wrapfigure}{r}{0.5\textwidth}
\begin{minipage}{0.5\textwidth}
\IncMargin{1.3em}
\vspace{-0.2in}
\begin{algorithm}[H]
\SetInd{0.5em}{0.8em}
\SetAlgoLined
\SetKwInput{KwInput}{Input}
\KwInput{``Posterior volume'' $\alpha$, number of samples $N$, optional burn-in $N_B$ and thinning period $N_T$. }
samples $\leftarrow [\,\,]$; \\
Get $\sigma$ from $\alpha$ by Eq.~\ref{eq:alpha-and-b-marginal} (matching) or \ref{eq:alpha-and-b-marginal-beta} (maximal); \\
Randomly initialize $t$; \\
\For{$i = 1, ..., N$}{
$t_{\mathrm{new}}, p_{\mathrm{for}}, p_{\mathrm{rev}} = \mathrm{propose}(t)$\\
Get $p$ from $t$ by Eq.~\ref{eq:bhat-and-relaxed-posterior} (match) or Eq.~\ref{eq:b-standardize-sigmoid-and-maximal-posterior} (max)\\
Get $p_{\mathrm{new}}$ from $t_\mathrm{new}$ by Eq.~\ref{eq:bhat-and-relaxed-posterior} or Eq.~\ref{eq:b-standardize-sigmoid-and-maximal-posterior};\\
$a \leftarrow (p_\mathrm{new} \cdot p_{\mathrm{rev}}) / (p\cdot p_{\mathrm{for}})$;\\
Sample $u\sim \mathcal{U}[0, 1]$;\\
\If{$u < a$}{
$t \leftarrow t_{\mathrm{new}}$;\\}
Append $t$ to samples; 
}
\vspace{0.02in}
Optionally, discard the first $N_B$ burn-in samples and thin the samples by only keeping every $N_T$ samples; \\
\Return{\em{samples}}\\
\caption{MH Sampling Procedure}
\label{alg:mh}
\end{algorithm}
\DecMargin{1.3em}
\end{minipage}
\end{wrapfigure}

The posterior sampling mechanism depends on the stochasticity of the controller and dynamics. 

\noindent \textbf{Deterministic Controller \& Dynamics}: 
When both the controller and the dynamics are deterministic, so is $\tau | t$, denoted as $\tau(t)$. Eq. \ref{eq:bhat-and-relaxed-posterior} reduces to 
    $t \sim p(t | \widehat b = b^*)\propto p(\widehat b=b^* | t, \tau(t))\pi(t), $
and similarly for Eq. \ref{eq:b-standardize-sigmoid-and-maximal-posterior}.

Alg.~\ref{alg:mh} presents the MH sampling procedure. First, $\sigma$ is computed from $\alpha$ (Line 2). Then we start with an initial task $t$ (Line 3). For each of the $N$ iterations, we propose a new task $t_\mathrm{new}$ according to a transition kernel and compute the forward and reverse transition probabilities $p_{\mathrm{for}}, p_{\mathrm{rev}}$ (Line 5). We evaluate the posteriors under $t$ and $t_\mathrm{new}$ (Line 6 and 7) and calculate the acceptance probability using the MH detailed balance principle (Line 8). Finally, we accept or reject accordingly (Line 9 -- 11). Note that if the proposal is rejected, the current $t$ is left unchanged \textit{and appended to the samples}. We can discard the first $N_B$ samples as burn-in, and/or thin the samples by a factor of $N_T$ to reduce auto-correlation.

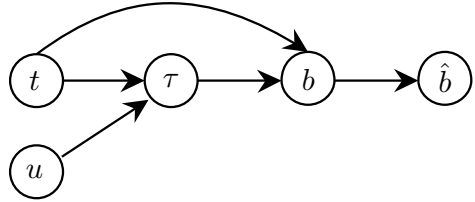
\begin{wrapfigure}{r}{0.5\textwidth}
    \centering
    \vspace{-0.4in}
    \resizebox{0.47\textwidth}{!}{
    \tikzset{every picture/.style={line width=0.75pt}} 

\begin{tikzpicture}[x=0.75pt,y=0.75pt,yscale=-1,xscale=1]

\draw   (156.31,68.59) .. controls (156.31,62.19) and (161.5,57) .. (167.91,57) .. controls (174.31,57) and (179.5,62.19) .. (179.5,68.59) .. controls (179.5,75) and (174.31,80.19) .. (167.91,80.19) .. controls (161.5,80.19) and (156.31,75) .. (156.31,68.59) -- cycle ;
\draw   (215.31,68.59) .. controls (215.31,62.19) and (220.5,57) .. (226.91,57) .. controls (233.31,57) and (238.5,62.19) .. (238.5,68.59) .. controls (238.5,75) and (233.31,80.19) .. (226.91,80.19) .. controls (220.5,80.19) and (215.31,75) .. (215.31,68.59) -- cycle ;

\draw   (275.31,68.59) .. controls (275.31,61.19) and (280.5,56) .. (286.91,56) .. controls (293.31,56) and (298.5,61.19) .. (298.5,68.59) .. controls (298.5,74) and (293.31,79.19) .. (286.91,79.19) .. controls (280.5,79.19) and (275.31,74) .. (275.31,68.59) -- cycle ;

\draw   (335.31,68.59) .. controls (335.31,61.19) and (340.5,56) .. (346.91,56) .. controls (353.31,56) and (358.5,61.19) .. (358.5,68.59) .. controls (358.5,74) and (353.31,79.19) .. (346.91,79.19) .. controls (340.5,79.19) and (335.31,74) .. (335.31,68.59) -- cycle ;
\draw    (179.5,68.59) -- (212.31,68.59) ;
\draw [shift={(215.31,68.59)}, rotate = 180] [fill={rgb, 255:red, 0; green, 0; blue, 0 }  ][line width=0.08]  [draw opacity=0] (10.72,-5.15) -- (0,0) -- (10.72,5.15) -- (7.12,0) -- cycle    ;
\draw    (238.5,68.59) -- (272.31,68.59) ;
\draw [shift={(275.31,68.59)}, rotate = 538.44] [fill={rgb, 255:red, 0; green, 0; blue, 0 }  ][line width=0.08]  [draw opacity=0] (10.72,-5.15) -- (0,0) -- (10.72,5.15) -- (7.12,0) -- cycle    ;
\draw    (298.5,68.59) -- (332.31,68.59) ;
\draw [shift={(335.31,68.59)}, rotate = 180] [fill={rgb, 255:red, 0; green, 0; blue, 0 }  ][line width=0.08]  [draw opacity=0] (10.72,-5.15) -- (0,0) -- (10.72,5.15) -- (7.12,0) -- cycle    ;
\draw    (167.91,57) .. controls (207.31,27.45) and (242.43,27.67) .. (284.96,54.74) ;
\draw [shift={(286.91,56)}, rotate = 213.12] [fill={rgb, 255:red, 0; green, 0; blue, 0 }  ][line width=0.08]  [draw opacity=0] (10.72,-5.15) -- (0,0) -- (10.72,5.15) -- (7.12,0) -- cycle    ;
\draw   (156.31,108.59) .. controls (156.31,102.19) and (161.5,97) .. (167.91,97) .. controls (174.31,97) and (179.5,102.19) .. (179.5,108.59) .. controls (179.5,115) and (174.31,120.19) .. (167.91,120.19) .. controls (161.5,120.19) and (156.31,115) .. (156.31,108.59) -- cycle ;

\draw    (179,102) -- (214.49,78.65) ;
\draw [shift={(217,77)}, rotate = 506.66] [fill={rgb, 255:red, 0; green, 0; blue, 0 }  ][line width=0.08]  [draw opacity=0] (10.72,-5.15) -- (0,0) -- (10.72,5.15) -- (7.12,0) -- cycle    ;

\draw (167,68.59) node  [align=center] {$\displaystyle t$};
\draw (227,68.59) node [inner sep=0.75pt]   [align=center] {$\displaystyle \tau $};
\draw (287,68) node [inner sep=0.75pt]   [align=center] {$\displaystyle b$};
\draw (347,67) node  [align=center] {$\displaystyle \hat{b}$};
\draw (167,108.59) node [align=center] {$\displaystyle u$};

\end{tikzpicture}
    }
    \caption{The same graphical model as in Fig.~\ref{fig:inference-diagram}, but with the addition of stochasticity $u$ in the controller such that $\tau | t, u$ is now deterministic.}
    \label{fig:inference-diagram-explicit}
\end{wrapfigure}

\noindent \textbf{Stochastic Controller}: 
When the controller and $p(\tau | t)$ are stochastic, the controller can usually be implemented by sampling a random variable $u$ (independent from $t$), and then producing the action based on the realization of $u$, as shown in Fig.~\ref{fig:inference-diagram-explicit}. For instance, a Normal stochastic policy $\pi(s)\sim \mathcal{N}(\mu(s), \sigma(s)^2)$ can be implemented by first sampling $u\sim \mathcal{N}(0, 1)$ and then computing $\pi(s) = \mu(s) + u\cdot \sigma(s)$. In this case, we sample in the combined $(t, \tau)$-space, with Eq.~\ref{eq:bhat-and-relaxed-posterior} being
$p(t, \tau | \widehat b=b^*) \propto p(\widehat b=b^* | t, \tau(e, u))p(u)\pi(t)$,
where we overload $\tau(t, u)$ to refer to the \textit{deterministic} trajectory given the task $t$ and controller randomness $u$. It is crucial that for any $u$, we can evaluate $p(u)$. Concretely, modifying Alg.~\ref{alg:mh}, $u_\mathrm{new}$ is proposed alongside with $t_\mathrm{new}$ (Line 5), the detailed balancing factor (Line 8) is multiplied by $p_{u, \mathrm{rev}} / p_{u, \mathrm{for}}$, and $t_\mathrm{new}, u_\mathrm{new}$ are accepted or rejected together (Line 10 -- 12). 

\noindent \textbf{Stochastic Dynamics}: Using the same logic, \modelname{} can also accommodate dynamics stochasticity, \textit{as long as it can be captured in a random variable $v$ and $p(v)$ can be evaluated.} We leave the details to App.~\ref{app:stochastic-dynamics} and use deterministic dynamics in our experiments.

\subsection{The Bayesian Posterior Sampling Interpretation}
\modelname{} uses Bayesian sampling concepts of prior, likelihood, and posterior quite liberally. Specifically, the task distribution is defined as the prior, and thus the notion of a task being likely in the deployment context refers to high probability under the prior. Likelihood refers to the behavior saliency: how much the exhibited behavior matches the behavior specification. The act of posterior sampling then finds tasks that strike a balance between these two objectives. 

The choice of explicitly modeling the task distribution is intentional, as it is not unlikely that the deployment environment will be different than the development environment. Such a domain mismatch may cause catastrophic failures, especially for learned controllers whose extrapolation behaviors are typically undefined. With a suitable task distribution, \modelname{} allows more failures to surface during this testing procedure.

\section{Behavior Taxonomy}
\label{sec:metrics}

Robot behaviors broadly belong to one of two classes: intentional and emergent. \textit{Intentional} behaviors are those that the controller explicitly optimize with objective functions. For example, the controller for a reaching task likely optimizes to move the end-effector to the target, by setting the target as an attractor in DS, using a target-reaching objective configuration in RRT, or rewarding proximity in RL. Thus, the final distance between the end-effector and the target is an intentional behavior for all three controllers. By contrast, \emph{emergent} behaviors are not explicitly specified in the objective. For the same reaching problem, an RL policy with reward based solely on distance may exhibit smooth trajectories for some target locations and jerky ones for others. Such behaviors may emerge due to robot kinematic structure, training stochasticity, or model inductive bias. 

For trajectory $\tau$, many behavior metrics $b(\tau, t)$ can be expressed as a line integral $\int_\tau V(\vec x)\d s$ of a scalar field $V(\vec x)$ along $\tau$ or its length-normalized version $\frac{1}{||\tau||}\int_\tau V(\vec x)\d s$, where $\d s$ is the infinitesimal segment on $\tau$ at $\vec x$ and $||\tau||$ is the trajectory length. $\vec x$ and $\tau$ can be in either joint space or task space. We introduce six behaviors: length, time derivatives (velocity, acceleration and jerk), straight-line deviation, obstacle clearance, near-obstacle velocity and motion legibility, whose mathematical expressions are in App.~\ref{app:behavior}. In addition, custom behaviors can also be used with \modelname{}.

\section{\modelname{} Use Case Demos}

In this section, we demonstrate how \modelname{} can find ``hidden'' properties of various controllers for two common tasks, navigation and reaching. We also uncover a suboptimal controller design due to bad hyper-parameter choices, which is improved based on \modelname{} insights. 

\subsection{Controller Algorithms}

\label{controller}
We consider four classes of robot controllers. The \textbf{imitation learning} (IL) controller uses expert demonstrations to learn a neural network policy which maps observations to deterministic actions. The \textbf{reinforcement learning} (RL) controller implements proximal policy gradient (PPO) \cite{schulman2017proximal}. While a mean and a variance is used to parameterize a PPO policy during training, the policy deterministically outputs the mean action during evaluation. The \textbf{dynamical system} (DS) controller modulates the linear controller $\vec u(\vec x)=\vec x^*-\vec x$, for the task-space target $\vec x^*$, into $\vec u_M(\vec x)=M\cdot \vec u(\vec x)$ using the modulation matrix $M$ derived from obstacle configuration, as proposed by \citet{huber2019avoidance}. We give a self-contained review in App.~\ref{app:ds-general}. The \textbf{rapidly-exploring random tree} (RRT) controller finds a configuration-space trajectory via RRT and then controls the robot through descretized segments. Notably, RRT is stochastic, and we discuss the use of controller stochasticity $u$ (c.f. Fig.~\ref{fig:inference-diagram-explicit}) in App.~\ref{app:rrt-general}.  
The MCMC sampling uses a Gaussian drift kernel, as detailed in App.~\ref{app:kernel}. 

\subsection{2D Navigation Task Experiments}

\begin{wrapfigure}{r}{0.55\textwidth}
    \centering
    \vspace{-0.25in}
    \includegraphics[width=0.55\columnwidth,trim={15px 8px 15px 20px},clip]{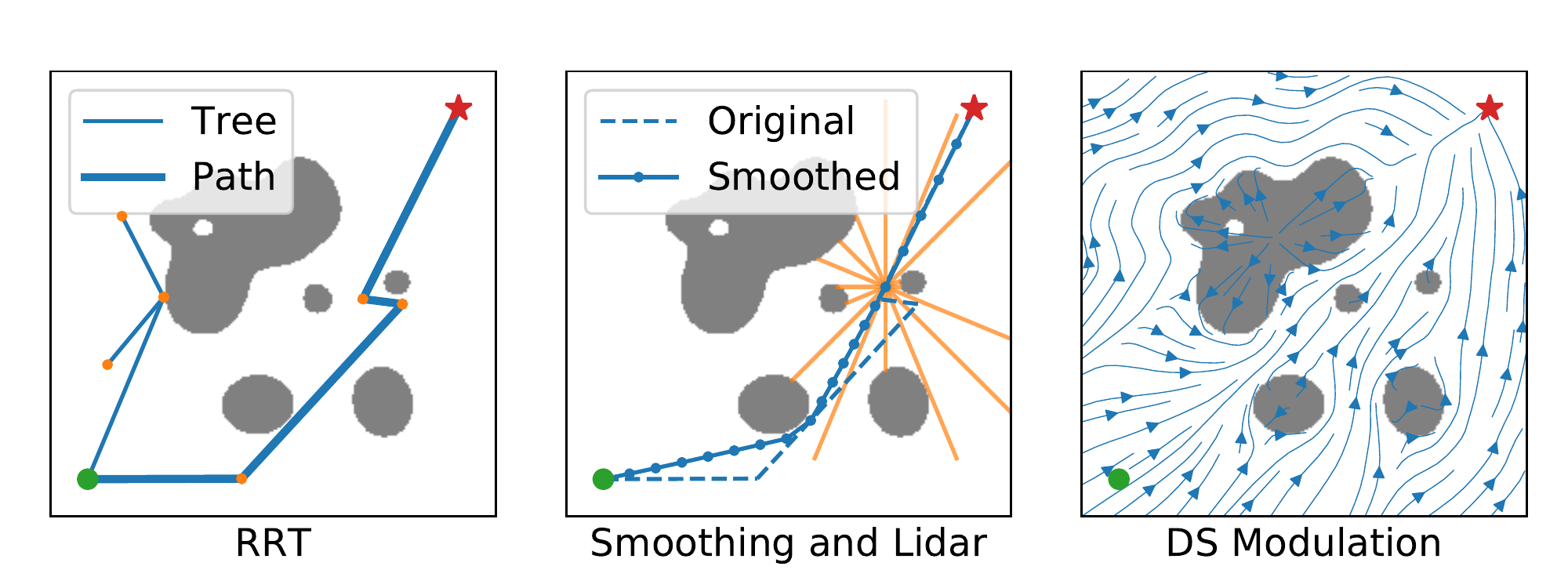}
    \vspace{-0.2in}
    \caption{RRT, IL and DS controllers on 2D navigation domain. Left: the RRT controller tree. Middle: smoothed RRT trajectory and lidar sensor (orange lines) for IL controller training. Right: the modulation by the DS controller. }
    \label{fig:2d_controllers}
\end{wrapfigure}

\textbf{Setup}\quad In a rectangular arena with irregularly shaped obstacles, a point mass robot needs to move from the lower left to the upper right corner (Fig.~\ref{fig:task-overview} left). App.~\ref{app:rbf-env-vis} details the obstacle generation and robot simulation procedures and contains more environment visualizations.

We consider three controllers for this environment: an RRT planner, a deep learning IL policy, and a DS (Fig.~\ref{fig:2d_controllers}). The RRT planner implements Alg.~\ref{alg:rrt} and discretizes the path to small segments as control signals at each time step. The IL controller uses smoothed RRT trajectories as expert demonstrations, and learns to predict heading angle from its current position and lidar readings. The DS controller finds an interior reference point for each obstacle, and converts each obstacle in the environment to be star-shaped. $\Gamma$-functions are then defined for these obstacles and used to compute the modulation matrix $M$. App.~\ref{app:2dnav-controller} contains additional implementation details. 

\textbf{Straight-Line Deviation}\quad In most cases, the robot cannot navigate directly to the target in a straight line. Thus, the collision-avoidance behavior is a crucial aspect for navigation robots. To understand it, we sample obstacles that lead to trajectories minimally deviating from the straight line path. Since the deviation is always non-negative, we use the matching mode in Eq. \ref{eq:bhat-and-relaxed-posterior} with target $b^*=0$. 

\begin{wrapfigure}{r}{0.5\textwidth}
    \centering
    \vspace{-0.03in}
    \includegraphics[width=0.49\columnwidth,trim={0px 150px 0px 0px},clip]{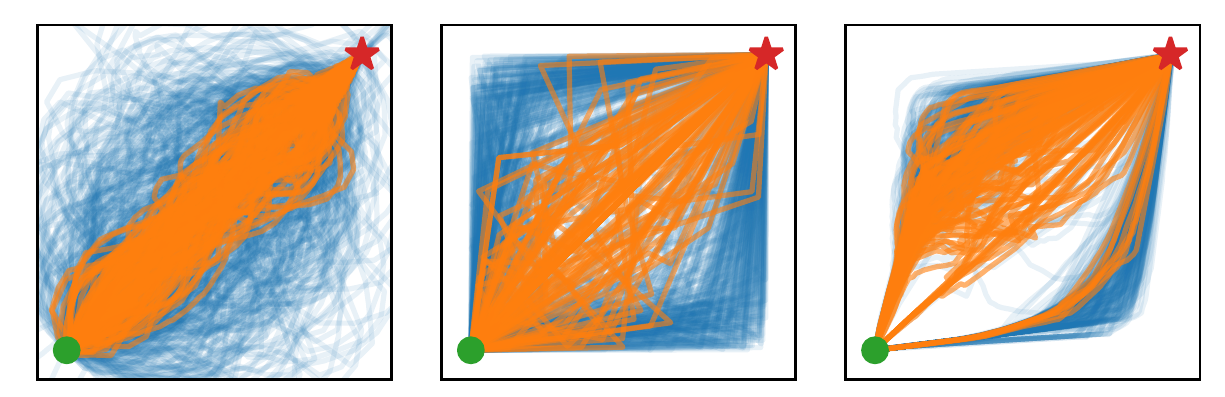}
    \includegraphics[width=0.49\columnwidth,trim={0px 5px 0px 127px},clip]{figures/2d_min_straight_line_deviation.pdf}
    \vspace{-2mm}
    \caption{Top: Posterior trajectories in orange vs.~prior in blue for minimal straight-line deviation behavior for three controllers. Bottom: Posterior obstacle distribution relative to the prior. Higher obstacle density regions are painted in red and lower ones in blue. }
    \label{fig:2d-min-center-deviation}
\end{wrapfigure}

In Fig.~\ref{fig:2d-min-center-deviation}, the top row plots posterior trajectories in orange, with prior trajectories in blue. The bottom row plots the obstacle distributions compared to the prior, with red regions being more likely to be occupied by obstacles and blue ones less likely to be obstructed.

For DS and RRT, the posterior trajectories and obstacle configurations are mostly symmetric with respect to the straight-line connection, as expected since both methods are formulated symmetrically with respect to the $x$- and $y$-coordinates. The obstacle distribution under RRT is also expected, since it seeks straight-line connections whenever possible and thus favor a ``diagonal corridor'' with obstacles on either side. For DS, however, obstacles are slightly \textit{more} likely to exist at the two ends of the above-mentioned corridor. This behavior is an artifact of the DS \emph{tail effect}, which drags the robot around the obstacle (details in App.~\ref{app:ds-general}). By taking advantage of anchor-like obstacles at the ends of the corridor, the modulation can reliably minimize the straight-line deviation. 

By comparison, the IL controller saliently exhibits trajectory asymmetry: it mostly takes paths on the left. It is possible that the asymmetry is due to ``unlucky'' samples by the MH sampler, but many independent restarts all confirm its presence, indicating that the asymmetry is inherent in the learned model. Since the neural network architecture is symmetric, we conclude that the stochasticity in the dataset generation and training procedure (e.g. initialization) leads to such imbalanced behaviors. Furthermore, the obstacle map suggests that obstacles are distributed very close to the robot path. Why does the robot seem to drive into obstacles? The answer lies in dataset generation: the smoothing procedure (Fig.~\ref{fig:2d_controllers} middle) results in most demonstrated paths navigating tightly around obstacles, and it is thus expected that the learned IL controller displays the same behavior. 

\textbf{Takeaways}\quad \modelname{} reveals two unexpected phenomena. First, IL trajectories are highly asymmetric toward the left of the obstacle due to dataset and/or training imbalance. Second, both DS and IL models exhibit certain ``obstacle-seeking'' behaviors, the former due to the ``tail-effect'' and the latter due the dataset generation process. In both cases, such behavior may be undesirable in deployment due to possibly imprecise actuation, and the controller design may need to be modified. 
Additional studies on legibility and obstacle clearance behaviors are presented in App.~\ref{app:2dnav-exp}.

\subsection{7DoF Arm Reaching Task Experiments}

\textbf{Setup}\quad A 7DoF Franka Panda arm is mounted on the side of a table with a T-shaped divider (Fig.~\ref{fig:task-overview} right). Starting from the same initial configuration on top of the table, it needs to reach a random location on either side under the divider. We simulate this task in PyBullet \cite{coumans2016pybullet}. We consider three controllers: an RRT planner, a deep RL PPO agent, and a DS formulation. 

RRT again implements Algorithm \ref{alg:rrt}, but uses inverse kinematics (IK) to first find the joint configuration corresponding to the target location. The RL controller is a multi-layer perceptron (MLP) network trained using the PPO algorithm. The DS model outputs the end-effector trajectory in the task space, which is converted to joint space via IK, with SVM-learned obstacle definitions.  App.~\ref{app:7dof-controller} contains additional implementation details for each method. Overall, RRT and RL are quite successful in reaching the target while the DS is not due to the bulky robot structure, close proximity to the divider, and the task-space only modulation. 

\begin{figure}[!htb]
    \centering
    \includegraphics[width=0.95\textwidth, trim=0 4.85in 0 0, clip]{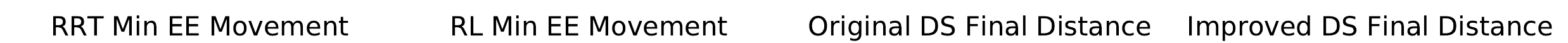}
    \includegraphics[width=0.95\textwidth, trim=0 2.8in 0 0.7in, clip]{figures/7dof_plots.pdf}
    \includegraphics[width=0.95\textwidth, trim=0 0.45in 0 3.05in, clip]{figures/7dof_plots.pdf}
    \vspace{-0.05in}
    \caption{Left: Minimal end-effector movement samples for RRT and RL. Right: Posterior samples for minimal distance from end-effector to target for the original and improved DS controllers. Top: posterior targets locations, with tabletop + divider in green and target region in orange. Bottom: posterior trajectories in red, prior trajectories in blue. Robot is mounted on the near long edge. 
    }
    \vspace{-0.2in}
    \label{fig:3d-min-ee-and-ds-improvement}
\end{figure}

\textbf{End-Effector Movement}\quad We find configurations that minimize the total travel distance of the end-effector for RRT and RL (DS omitted due to high failure rate). Fig.~\ref{fig:3d-min-ee-and-ds-improvement} (left two) shows the posterior target locations and trajectories. Notably, unlike RL, RRT trajectories are highly asymmetric, since there are straight-line connections in the configuration space from the initial pose to some target regions on the left, while every right-side goal requires at least an intermediate node. 

\textbf{DS Improvement with \modelname{}}\quad
Our initial DS implementation frequently fails to reach the target. This is understandable, as the DS convergence guarantee~\cite{huber2019avoidance} is only valid in task space, in which the modulation is defined. When the full-arm motion is solved via IK, it is possible that some body parts may collide and get stuck because of the table divider. 
To understand the DS behaviors, we use \modelname{} to sample target locations that result in minimal final distance from the end-effector to the target (i.e., most successful executions,  Fig.~\ref{fig:3d-min-ee-and-ds-improvement} center-right). Similar to the RRT case, the samples show strong lateral asymmetry, with all posterior target locations on the left, due to the same cause of asymmetric kinematic structure. 
The result points to a clear path to improve the DS controller such that it can succeed with right-side targets: increase the collision clearance of the divider so that the end-effector navigates farther away from the divider, thus also bringing the whole arm to be farther away. As detailed in App.~\ref{app:ds-improvement}, this modification greatly improves the controller performance as confirmed by the new symmetry in Fig.~\ref{fig:3d-min-ee-and-ds-improvement} (rightmost). In addition, since the issue with DS controller mainly lies in obstacle avoidance in joint-space or on the body of the robot, additional techniques \citep{khatib1986real, ratliff2018riemannian, mirrazavi2018unified, urain2021composable} could be used and we leave them to future directions. 

\textbf{Takeaway}\quad The set of studies reveal an important implication of the robot's kinematic structure: the left side is much less ``congested'' with obstacles than the right side in the configuration space. While the RL controller is able to learn efficient policies for both sides, the design of certain controllers may need to explicitly consider such factors. App. \ref{app:3d-additional-vis} includes an additional study on legibility. 

\subsection{Quantitative Summary}
We studied other additional behaviors on both tasks, and Tab.~\ref{tab:quant-conf-main} summarizes prior vs. posterior mean behavior values and shows that \modelname{} consistently finds samples salient in the target behavior. 

\begin{table}[!htb]
\centering
\vspace{0.1in}
\resizebox{\textwidth}{!}{
\begin{tabular}{ lllrrrrrr }
\toprule
Domain & Behavior & Target & Prior (DS) & Post. (DS) & Prior (IL/RL) & Post. (IL/RL) & Prior (RRT) & Post. (RRT) \\  \midrule
\multirow{6}{*}{2D Nav}
                    &   Avg.~Jerk      & 0 & 1.84e-3 & 1.46e-3 & 6.95e-4 & 3.19e-4 & 4.24e-4 & 2.79e-4 \\
                    &   Straight       & 0 & 0.256 & 0.084 & 0.378 & 0.301 & 0.470 & 0.162\\
                    &   Legibility     & min & 0.819 & 0.650 & 0.877 & 0.784 & 0.798 & 0.669\\
                    &   Obstacle       & 0 & 0.309 & 0.229 & 0.262 &  0.218 & 0.312 & 0.241\\
                    &   Obstacle       & max & 0.309 & 0.611 & 0.262 & 0.387 & 0.312 & 0.442\\\midrule
\multirow{2}*{Arm} & Straight & 0 & 0.980 & 0.913 & 0.858 & 0.762 & 1.223 & 0.897\\
                   & EE Dist  & 0 & 0.934 & 0.623 & 0.958 & 0.691 & 3.741 & 1.192\\\bottomrule
\end{tabular}
}
\vspace{0.01in}
\caption{Quantitative results on additional tasks for two domains. }
\label{tab:quant-conf-main}
\end{table}

\section{MCMC Sampling Evaluation}

\begin{wrapfigure}{r}{0.35\textwidth}
    \centering
    \vspace{-0.28in}
    \includegraphics[width=0.35\textwidth]{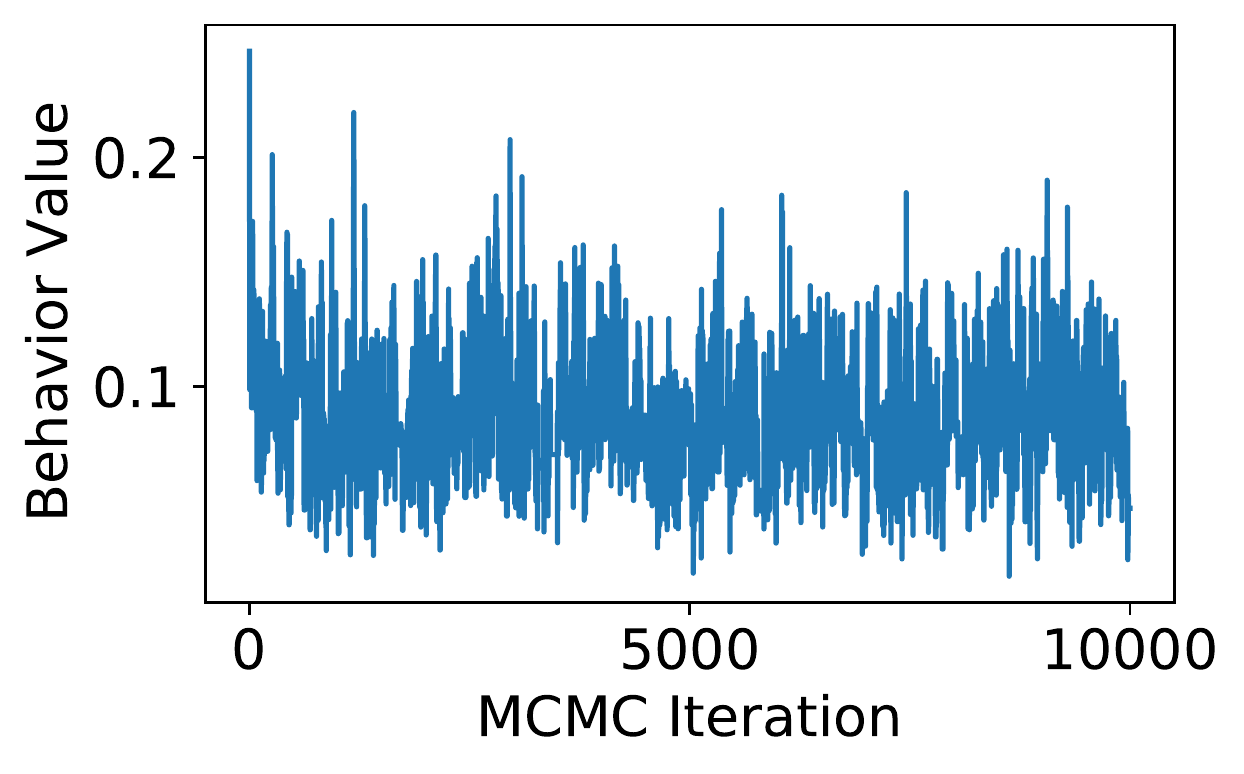}
    \vspace{-0.25in}
    \caption{2D navigation DS min straight-line deviation samples.}
    \vspace{-0.04in}
    \label{fig:num-samples}
\end{wrapfigure}

After confirming that \modelname{} can indeed uncover significant and actionable controller insights, we evaluate the sampling procedure itself, using tasks described above as examples.

\textbf{Mixing Property} \quad A potential downside of MCMC sampler is the slow mixing time, which causes the chain to take a long time to converge from initialization and causes consecutive samples to be highly correlated. Does this phenomenon happen for our sampling? Fig.~\ref{fig:num-samples} plots the behavior along the MCMC iterations for the DS minimal straight-line deviation behavior, showing that the chain mixes well quite fast (additional ones in Fig.~\ref{fig:num-samples-app} of App.~\ref{app:kernel}). Thus, a modest amount of samples, such as several thousand, is typically sufficient to model the target posterior distribution well.

\textbf{Baseline: Top-$k$ Selection} \quad To the best of our knowledge, \modelname{} is the first work that applies the transparency-by-example formulation \citep{booth2020bayes} to robotic tasks, and we are not aware of existing methods for the same purpose. Notably, adversarial perturbation algorithms \citep{goodfellow2014explaining} are \textit{not} feasible, since stepping in simulator (or real world) is not typically differentiable. Sec.~\ref{sec:intro} discusses a straightforward alternative that runs the controller on $N$ different scenarios and pick the top-$k$ with respect to the target behavior. We demonstrate its shortcomings on the minimal straight-line deviation behavior for the 2D navigation DS controller (\modelname{} samples shown in Fig.~\ref{fig:2d-min-center-deviation} left). 

\begin{figure}[!htb]
    \centering
    \vspace{0.1in}
    \includegraphics[width=\textwidth]{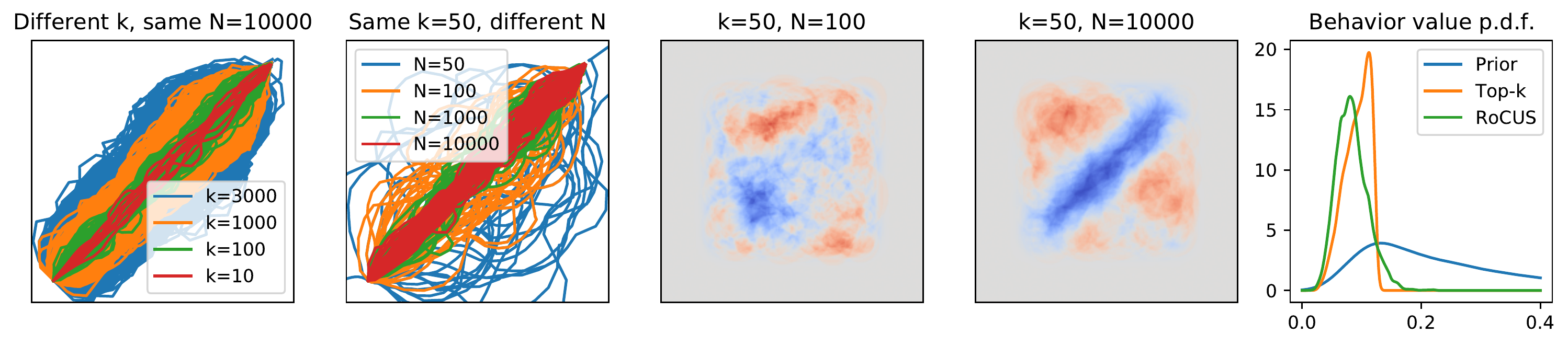}
    \vspace{-0.3in}
    \caption{Top-$k$ selection baseline. Left two: trajectory distribution; middle two: obstacle distribution; right one: probability density function of behavior values. }
    \label{fig:top-k-experiment}
\end{figure}

Fig.~\ref{fig:top-k-experiment} (left) shows the trajectories of different values of $k$ for the same fixed $N$, and vice versa. While a bigger $N/k$ ratio leads to more salient behaviors in the top-$k$ samples, these examples become more concentrated around the global maximum and less diverse, making this approach especially myopic. Further, it is not easy to find the optimal $N$ to trade off between diversity and saliency of the top-$k$ samples. By contrast, \modelname{} offers the intuitive $\alpha$ hyper-parameter. Fig.~\ref{fig:top-k-experiment} (middle) shows that a smaller $N$ fails to highlight the ``corridor'' pattern while a larger $N$ makes it completely open and misses the ``tail-effect anchors'' at the two ends. 

In addition, the hard cut-off at the $k$-th salient behavior threshold has two undesirable implications: first, every trajectory more salient than the threshold is kept but is given equal importance; second, a trajectory even slightly under the threshold is strictly discarded. By comparison, \modelname{} gives more importance to more salient samples in a progressive manner, as shown in Fig.~\ref{fig:top-k-experiment} right. 

Finally, top-$k$ selection is very computationally inefficient. It discards all of the unselected $N-k$ samples, while \modelname{} is much more efficient in that all samples after the burn-in up to the thinning factor can be kept since the posterior concentrated on the salient behavior is directly sampled. 

\section{Discussion and Future Work}
\label{sec:conclusion}

\modelname{} enables humans to build better mental models of robot controllers. Compared to existing  evaluations on task-completion metrics for hand-designed tasks, \modelname{} generates tasks and trajectories that highlight any given behavior in a principled way. We used it to uncover non-obvious insights in two domains and help with debugging and improving a controller. 

While \modelname{} is mainly a tool to analyze robot controllers in simulation as part of comprehensive testing before deployment, it can help understanding (anomalous) real world behaviors as well. When an anomaly is observed, \modelname{} can find more samples with the anomaly for developers to identify patterns of systematic failures. Furthermore, \modelname{} is not inherently limited to simulation: it only requires trajectory roll-out on specific tasks. For the arm reaching task, this is easy in the real world. For autonomous driving, ``recreating'' a traffic condition that involves other vehicles may be hard. However, a key feature of \modelname{} is the decoupling of the task and the controller algorithm, which allows testing on simpler task variants (e.g. with props instead of real cars).

There are multiple directions for future work, including evaluation of \textit{model updates}~\citep{bansal2019updates} by defining behavior functions on two controllers, better understanding the samples with explainable artificial intelligence (XAI) methods, and an appropriate interface to facilitate the two-way communication between \modelname{} and end-users, as discussed in detail in App.~\ref{app:future-work}. 

Overall, \modelname{} is a framework for systematic discovery and inspection of robotic controller behaviors. We hope that the demonstrated utility of \modelname{} sparks further efforts towards the development of other tools for more holistic understanding of robot controllers.

\clearpage

\section*{Acknowledgement}
This research is supported by the National Science Foundation (NSF) under the grant IIS-1830282. We thank the reviewers for their reviews, which are available at \url{https://openreview.net/forum?id=5P_3bRWiRsF}.

\bibliography{references_corl}

\clearpage

\appendix

\section{Scale-Invariance and ``Volume'' Interpretation of $\alpha$}
\label{supp:scale-invariance}

We show that Eq. \ref{eq:alpha-and-b-marginal} results in the formulation being scale-invariant with respect to $b$. Consider the same behavior under two different units $b_1$ and $b_2$ with $b_1=c\cdot b_2$. For example, $b_1$ can be the trajectory length in centimeters and $b_2$ is the same quantity but in meters, and $c=100$. Thus, $p(c\cdot b_1) = p(b_2)$ and $b_1^*=c\cdot b_2^*$. To maintain the same $\alpha$ level in Eq. \ref{eq:alpha-and-b-marginal}, we need to have $\sigma_1=c\cdot \sigma_2$. This implies that
\begin{align}
&p(t, \tau | \hat b_1=b_1^*) = \frac{\mathcal{N}(b_1^*; b(\tau, t), \sigma_1^2) p(\tau|t)\pi(t)}{p(\hat b_1=b_1^*)}\\
=&\frac{\mathcal{N}(b_2^*; b(\tau, t), \sigma_2^2) p(\tau|e)\pi(t)}{p(\hat b_2=b_2^*)}=p(t | \hat b_2=b_2^*)
\end{align}
because $\mathcal{N}(b_1^*; b(\tau, t), \sigma_1^2)=\mathcal{N}(b_2^*; b(\tau, t), \sigma_2^2)$ due to the same scaling of $b_1 \sim b_2$ and $\sigma_1 \sim \sigma_2$, and $p(\hat b_1=b_1^*)=p(\hat b_2=b_2^*)$ as they are the same event. We conclude that the posterior distribution is scale-invariant with respect to $b(\tau, t)$. 

To motivate the bound of $[b^* - \sqrt{3}\sigma, b^* + \sqrt{3}\sigma]$ in Eq. \ref{eq:alpha-and-b-marginal}, we consider a uniform approximation to $\mathcal{N}(b^*, \sigma^2)$. To match the mean $b^*$ and standard deviation $\sigma$, $\mathcal{U}(b^*-\sqrt{3}\sigma, b^*+\sqrt{3}\sigma)$ is needed. If we use this uniform distribution in Eq. \ref{eq:bhat-and-relaxed-posterior} in lieu of the normal distribution, the posterior can be instantiated by sampling from the prior and rejecting tasks for which the trajectory behavior $b(\tau, t)$ falls outside of this bound. Thus, Eq. \ref{eq:alpha-and-b-marginal} specifies that the ``volume'' of $(\alpha\cdot 100)\%$ under $p(t, \tau)$ is maintained.

The same invariance and ``volume'' interpretation holds for Eq. \ref{eq:alpha-and-b-marginal-beta} as well. The former stems from the standardization on $b$ performed in Eq. \ref{eq:b-standardize-sigmoid-and-maximal-posterior}. The latter uses the same uniform approximation but the bound is one-sided since $\beta\in(0, 1)$ by nature of the sigmoid transformation. 

\section{MCMC Sampling with Stochastic Dynamics}
\label{app:stochastic-dynamics}
Using the same logic as the case of stochastic controller, \modelname{} can also accommodate stochasticity in transition dynamics (e.g. object position uncertainty after it is pushed), \textit{as long as such stochasticity can be captured in a random variable $v$ and $p(v | t)$ can be evaluated.} This is typically possible in simulation, and the modification to Alg.~\ref{alg:mh} is similar to the case of stochastic controllers. In the real world, we can 
\begin{itemize}[leftmargin=0.2in, topsep=-3pt, itemsep=0pt, parsep=0pt]
    \item treat a sampled trajectory as the deterministic one; 
    \item restart multiple times to estimate $\mathbb E_\tau[b(\tau, t)]$; or
    \item use likelihood-free MCMC methods \cite{brooks2011handbook}. 
\end{itemize}
We leave these investigations to future work, and use deterministic dynamics in our experiments.

\section{Mathematical Definitions of Behaviors}
\label{app:behavior}
A versatile and general form of a behavior is the (normalized or unnormalized) line integral of some scalar field along the trajectory. Specifically, we have

\begin{align}
    b = \int_\tau V(\vec x)\d s \quad \mathrm{or}\quad b = \frac{1}{||\tau||}\int_\tau V(\vec x)\d s.
\end{align}

Using this general definition, we define a list of behaviors in Tab.~\ref{tab:behavior-list}.

\begin{table}[!htb]
    \centering
    \vspace{0.1in}
    \resizebox{\textwidth}{!}{
    \begin{tabular}{ll|ll}\toprule 
        Name & Definition & Name & Definition\\\midrule
        Trajectory Length & $\displaystyle b = \int_\tau 1 \d s$ & Straight-Line Deviation & $\displaystyle b = \frac{1}{||\tau||}\int_\tau ||\vec x - \mathrm{proj}_{\vec x_f - \vec x_i} \vec x|| \d s$\\[2ex]
        Average Velocity & $\displaystyle b = \frac{1}{||\tau||}\int_\tau ||\dot{\vec x}|| \d s$ & Obstacle Clearance & $\displaystyle b=\frac{1}{||\tau||}\int_\tau \min_{\vec x_o\in \mathcal O}||\vec x - \vec x_o||\d s$\\[2.5ex]
        Average Acceleration & $\displaystyle b = \frac{1}{||\tau||}\int_\tau ||\ddot{\vec x}|| \d s$ & Near-Obstacle Velocity & $\displaystyle b = \frac{\int_\tau ||\dot{\vec x}|| / \min_{\vec x_o\in \mathcal O}||\vec x - \vec x_o|| \d s}{\int_\tau 1 / \min_{\vec x_o\in \mathcal O}||\vec x - \vec x_o|| \d s}$\\[2.5ex]
        Average Jerk & $\displaystyle b = \frac{1}{||\tau||}\int_\tau ||\dddot{\vec x}|| \d s$ & Motion Legibility & $\displaystyle b = \frac{1}{||\tau||}\int_\tau p(g|\vec x) \d s$\\[1.8ex]
        \bottomrule
    \end{tabular}
    }
    \vspace{0.05in}
    \caption{A list of behavior definitions. }
    \label{tab:behavior-list}
\end{table}

\textbf{Trajectory length} simply measures how long the trajectory is. In most of the behaviors below, the normalizing factor is also length to decorrelate the behavior value from it. 

\textbf{Average velocity, acceleration and jerk} are useful for a general understanding about how fast and abruptly the robot moves, which is an important factor to its safety. 

\textbf{Straight-line deviation} measures how much the robot trajectory deviates from the straight-line path, in either the task space or the state space. A specific task instance in which the straight-line path is feasible (e.g. with no obstacles) is typically considered easy. Thus, we can find tasks of varying difficulty level on the spectrum of deviation values. In the definition, $\vec x_i$ is the initial state, $\vec x_f$ is the final state, and $\mathrm{proj}$ is the projection operator. 

\textbf{Obstacle clearance} measures the average distance to the closest obstacle. Finding situations in which the robot moves very close to obstacles is crucial to understanding the collision risk level. In the definition, $\mathcal O$ represents the obstacle space. 

\textbf{Near-obstacle velocity} calculates how fast the robot moves around obstacles. We define it as the average velocity on the trajectory weighted by the inverse distance to the closest obstacle. Other weighting method can be used, as long as it is non-negative and monotonically decreasing with distance. This behavior is correlated with the damage of a potential collision, as high-speed collisions are usually far more dangerous and costly. Since we want the value to represent the average velocity, we normalize by the integral of weights along the trajectory. 

\textbf{Motion legibility} measures how well the goal can be predicted over the course of the exhibited trajectory. In our definition, we use $p(g|\vec x)$, or the conditional probability of the goal $g$ given at the current robot state $\vec x$, but there may be better application-specific definitions.

\section{Dynamical System Modulation}
\label{app:ds-general}

We review the DS formulation proposed by \citet{huber2019avoidance}, and present our problem-specific adaptations for 2D Navigation in App.~\ref{app:2d-ds} and 7DoF arm reaching in App.~\ref{app:3d-ds}. A reader familiar with DS motion controllers may skip this review.

Given a target $\vec x^*$ and the robot's current state $\vec x$, a linear controller $\vec u(x)=\vec x^* - \vec x$ will guarantee convergence of $\vec x$ to $\vec x^*$ if there are no obstacles. However, it can easily get stuck in the presence of obstacles. \citet{huber2019avoidance} proposes a method to calculate a modulation matrix $M(\vec x)$ at every $\vec x$ such that if the new controller follows $\vec u_M(\vec x)=M(\vec x)\cdot \vec u(\vec x)$, then $\vec x$ still converges to $\vec x^*$ but never gets stuck, as long as $\vec x^*$ is in free space. In short, the objective of the DS modulation is to preserve the linear controller's convergence guarantee while also ensuring that the robot is never in collision.

The modulation matrix $M(\vec x)$ is computed from a list of obstacles, each of which is represented by a $\Gamma$-function. For the $i$-th obstacle $\mathcal O_i$, its associated gamma function $\Gamma_{i}$ must satisfy the following properties: 
\begin{itemize}[leftmargin=0.2in, topsep=0.0in, itemsep=0.0in]
    \item $\Gamma_i(\vec x)\leq 1 \iff \vec x \in \mathcal O_i$, 
    \item $\Gamma_i(\vec x) = 1 \iff \vec x \in \partial \mathcal O_i$, 
    \item $\exists \,\vec r_i, \mathrm{s.t.} \forall \, t_1\geq t_2\geq 0, \forall \, \vec u, \Gamma_i(\vec r_i + t_1\vec u) \geq \Gamma_i(\vec r_i + t_2\vec u)$. 
\end{itemize}
In words, the $\Gamma$-function value needs to be less than 1 when inside the obstacle, equal to 1 on the boundary, greater than 1 when outside. This function must also be monotonically increasing radially outward from a specific point $\vec r_i$. This point is dubbed the \emph{reference point}. From this formulation, $\vec r_i\in \mathcal O_i$ and any ray from $\vec r_i$ intersects with the obstacle boundary $\partial \mathcal O_i$ exactly once. The latter property is also the definition that $\mathcal O_i$ is ``star-shaped'' (Fig.~\ref{fig:star-shaped-explainer-supp}). For most common (2D) geometric shape such as rectangles, circles, ellipses, regular polygons and regular stars, $\vec r_i$ can be chosen as the geometric center. 

We first consider the case of a single obstacle $\mathcal O$, represented by $\Gamma$ with reference point $\vec r$. Use $d$ to denote the dimension of the space. We define 
\begin{align}
    M(\vec x)=E(\vec x)D(\vec x)E^{-1}(\vec x). 
\end{align} 
We have 
\begin{align}
E(x)=[\vec s(\vec x), \allowbreak \vec e_1(\vec x), \allowbreak ..., \allowbreak \vec e_{d-1}(\vec x)], 
\end{align}
where 
\begin{align}
    \vec s(\vec x) = \frac{\vec x - \vec r}{||\vec x - \vec r||}
\end{align}
is the unit vector in the direction of $\vec x$ from $\vec r$, and $\vec e_1(\vec x), \allowbreak ..., \allowbreak \vec e_{d-1}(\vec x)$ form a $d-1$ orthonormal basis to the gradient of the $\Gamma$-function, $\nabla\Gamma(\vec x)$ representing the normal to the obstacle
surface. $D(\vec x)$ is a diagonal matrix whose diagonal entries are $\lambda_s, \lambda_1, ..., \lambda_{d-1}$, with
\begin{align}
    \lambda_s &= 1 - \frac{1}{\Gamma(\vec x)}, \\
    \lambda_1, ..., \lambda_{d-1} &= 1 + \frac{1}{\Gamma(\vec x)}. 
\end{align}
each eigenvalue determines the scaling of each direction. Conceptually, as the robot approaches the obstacle, this modulation decreases the velocity for the component in the reference point direction (i.e. toward obstacles) while increases velocity for perpendicular components. The combined effect results in the robot being deflected away tangent to the obstacle surface.

With $N$ obstacles, we compute the modulation matrix $M_i(\vec x)$ for every obstacle using the procedure above and the individual controllers $\vec u_{M_i}(\vec x)=M_i(\vec x)\cdot \vec u(\vec x)$. The final modulation is the aggregate of all the individual modulations. However, a simple average is insufficient since closer obstacles should have higher influence to prevent collisions.

\citet{huber2019avoidance} proposed the following aggregation procedure. Let  $\vec u_i$ denote the individual modulations, with norms $n_i$. The final aggregate modulation $\vec u$ is calculated as
\begin{align}
    \vec u = n_a \vec u_a, \label{final aggregation}
\end{align}
where $n_a$ and $\vec u_a$ are the aggregate norm and direction. 

The aggregate norm is computed as
\begin{align}
    n_a &= \sum_{i=1}^N w_i n_i, \\
    w_i &= \frac{b_i}{\sum_{j=1}^N b_j}, \\
    b_i &= \prod_{1\leq j\leq N, j\neq i}\Gamma_{j}(\vec x). 
\end{align}
The above definition ensures that $\sum_{i=1}^N w_i=1$, and $w_i\rightarrow 1$ when $\vec x$ approaches $\mathcal O_i$ (and only $\mathcal O_i$, which holds as long as obstacles are disjoint). 

$\vec u_a$ is instead computed using what \citet{huber2019avoidance} calls ``$\boldsymbol \kappa$-space interpolation.'' First, similar to the basis vector matrix $E(\vec x)$ introduced above, we construct another such matrix, but with respect to the original controller $\vec x^* - \vec x$. We denote it as $R=[(\vec x^* - \vec x) / ||\vec x^* - \vec x||, \vec e_1, ..., \vec e_{d-1}]$, where $\vec e_1, ..., \vec e_{d-1}$ are again orthonomal vectors spanning the null space. 

For each $\vec u_i$, we compute its coordinate in this new $R$-frame as $\hat {\vec u}_i=R^{-1}\vec u_i$. Its $\boldsymbol \kappa$-space representation is
\begin{align}
    \boldsymbol \kappa_i =\frac{\arccos(\hat {\vec u}_i^{(1)})}{\sum_{m=2}^d \hat {\vec u}_i^{(m)}} \left[\hat {\vec u}_i^{(2)}, ..., \hat {\vec u}_i^{(d)}\right]^T \in \mathbb R^{d-1}, 
\end{align}
where the superscript $(m)$ refers to the $m$-th entry. $\boldsymbol \kappa_i$ is a scaled version of the $\hat{\vec u}_i$ with the first entry removed. We perform the aggregation in this $\boldsymbol \kappa$-space using the weights $w_i$ calculated above (\ref{weights_agg}), transform it back to the $R$-frame (\ref{eq:ref-frame-transform}), and finally transform it back to the original frame (\ref{eq:original_frame_transform}): 
\begin{align}
    \boldsymbol \kappa_a &= \sum_{i=1}^N w_i \boldsymbol \kappa_i \label{weights_agg} \\
    \hat {\vec u}_a &= \left[\cos(||\boldsymbol \kappa_a||), \frac{\sin(||\boldsymbol \kappa_a||)}{||\boldsymbol \kappa_a||} \boldsymbol \kappa_a ^T\right]^T \label{eq:ref-frame-transform}\\
    \vec u_a &= R\hat {\vec u}_a.\label{eq:original_frame_transform}
\end{align}
As mentioned in Eq. \ref{final aggregation}, the final modulation is $\vec u = n_a \vec u_a$. 

\subsection{Tail-Effect}
\begin{figure}[!htb]
    \centering
    \vspace{0.1in}
    \includegraphics[width=0.6\columnwidth]{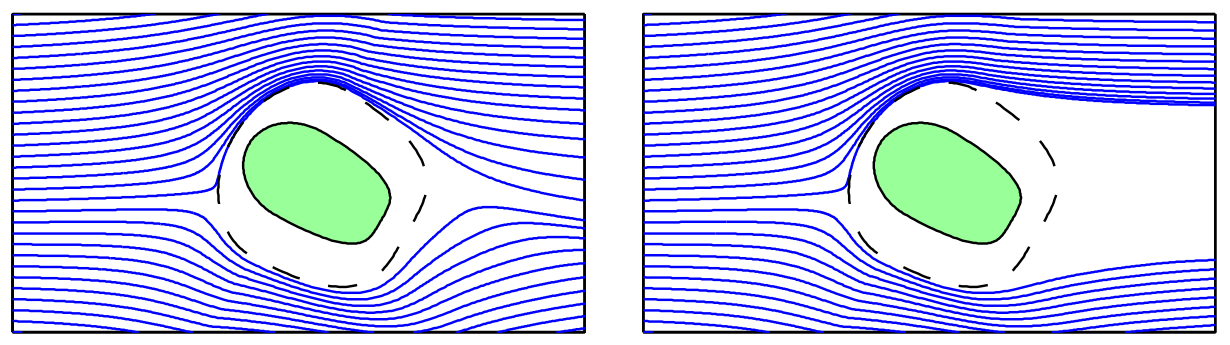}
    \caption{Tail effect (left) and its removal (right), reproduced from Fig.~7 by \citet{khansari2012dynamical}. The target is on the far right side. }
    \label{fig:tail-effect}
\end{figure}

An artifact of the above formulation is the ``tail-effect,'' where the robot is modulated to go around the obstacle even when it has passed by the obstacle and the remaining trajectory has no chance of collision under the non-modulated controller. This effect has been observed by \citet{khansari2012dynamical} for a related but different type of modulation. Fig. \ref{fig:tail-effect}, reproduced from the paper by \citet[][Fig.~7]{khansari2012dynamical}, shows the tail effect on the left and its removal on the right. This tail effect induces the placement of obstacles at the end of the ``diagonal corridor'' as seen in our straight-line deviation experiments (Fig.~\ref{fig:2d-min-center-deviation}, left). If desired, the DS formulation can be modified to remove this effect.

\section{RRT Algorithm Description and Sampling}
\label{app:rrt-general}
There are many RRT variants with subtle differences. For clarity, Algorithm \ref{alg:rrt} presents the version that we use. 

\IncMargin{1.3em}
\vspace{0.1in}
\begin{algorithm}[!htb]
\SetInd{0.5em}{0.8em}
\SetAlgoLined
\SetKwInput{KwInput}{Input}
\KwInput{Start configuration $s_0$, target configuration $s^*$. }
$\mathcal{T}$ $\leftarrow$ tree(root = $s_0$);\\
success $\leftarrow$ attempt-grow($\mathcal{T}$, from = $s_0$, to = $s^*$); \\
\While{not \em{success}}{
$s$ $\leftarrow$ sample-configuration($\,$); \\
$s_n$ $\leftarrow$ nearest-neighbor($\mathcal{T}$, $s$);\\
success $\leftarrow$ attempt-grow($\mathcal{T}$, from = $s_n$, to = $s$); \\
\If{\em{success}}{
success $\leftarrow$ attempt-grow($\mathcal{T}$, from = $s$, to = $s^*$); \\
}
}
\vspace{0.02in}
\Return{\em{path(}$\mathcal{T}$, from = $s_0$, to = $s^*$)}
\caption{RRT Algorithm}
\label{alg:rrt}
\end{algorithm}
\DecMargin{1.3em}

While RRT is stochastic (unlike DS, IL and RL), the entire randomness is captured by the sequence of C-space samples used to grow the tree, including failed ones. We call this a \textit{growth} $g=[s_1, s_2, s_3, ...]$. The probabilistic completeness property of RRT generally assures that the algorithm will terminate in finite time with probability 1 if a path to the target exists \cite{lavalle2006planning}. 
Thus, hypothetically, given an infinitely long tape containing every entry of $g$, we can compute a deterministic trajectory $\tau=\mathrm{RRT}(s_0, s^*, g)$ with a finite number of nodes with probability 1. 

To enable MH inference, we take inspiration from Bayesian nonparametrics: we instantiate $g$ on an \textit{as-needed} basis. We start with an empty vector of $g=[\,]$. When calculating $\mathrm{RRT}(s_0, s^*, g)$, if a new point beyond existing entries of $g$ needs to be sampled, we append it to $g$. During MH inference, we use a transition kernel that operates element-wise on instantiated entries of $g$ (i.e. independently perturbing each entry of $g$). If the transition kernel does not depend on the current $g$ (e.g. drawing uniformly from the C-space), then past instantiated entries do not even need to be kept. 

Note that RRT trajectories are often smoothed \textit{post hoc}. Since our main focus is to evaluate and identify problems for an existing one, we use the original formulation. Moreover, it is easy to use \modelname{} to evaluate model updates (e.g. original vs smoothed RRT) as discussed in Sec.~\ref{sec:conclusion}.

\section{MCMC Sampling Details}
\label{app:kernel}

We used a truncated Gaussian transition kernel for all experiments. For the RBF-defined 2D environment, we initialize 15 obstacle points with coordinates sampled uniformly in $[-0.7,0.7]$. The transition kernel operates independently on each obstacle coordinate: given the current value of $x$, the kernel samples a proposal from $\mathcal{N}(\mu=x, \sigma^2=0.1^2)$ truncated to $[-0.7, 0.7]$ (and also appropriately scaled). For the arm reaching task, the target is sampled uniformly from two disjoint boxes, with the left box at $[-0.5, -0.05]\times[-0.3, 0.2]\times[0.65, 1.0]$ and the right box at $[0.05, 0.5]\times[-0.3, 0.2]\times[0.65, 1.0]$. Again, we use the same transition kernel with $\sigma_x=0.1, \sigma_y=0.03, \sigma_z=0.035$ in three directions. Again, the distribution is truncated to the valid target region ($x\in[-0.5, -0.05]\cup[0.05, 0.5], y\in[-0.3, 0.2], z\in[0.65, 1.0]$). In other words, the transition kernel implicitly allows for the jump across two box regions. 

In addition, the stochastic RRT controller also requires a transition kernel. As discussed in Sec. \ref{controller}, we initialize its values on an as-needed basis. When necessary, we sample a configuration uniformly between the lower- and upper-limit (i.e. $[x_L, x_U]$). For each configuration, the same Gaussian kernel truncated to $[x_L, x_U]$, and $\sigma=0.1(x_U-x_L)$ is used. 

\begin{figure}[!htb]
    \centering
    \vspace{0.1in}
    \includegraphics[width=0.7\textwidth]{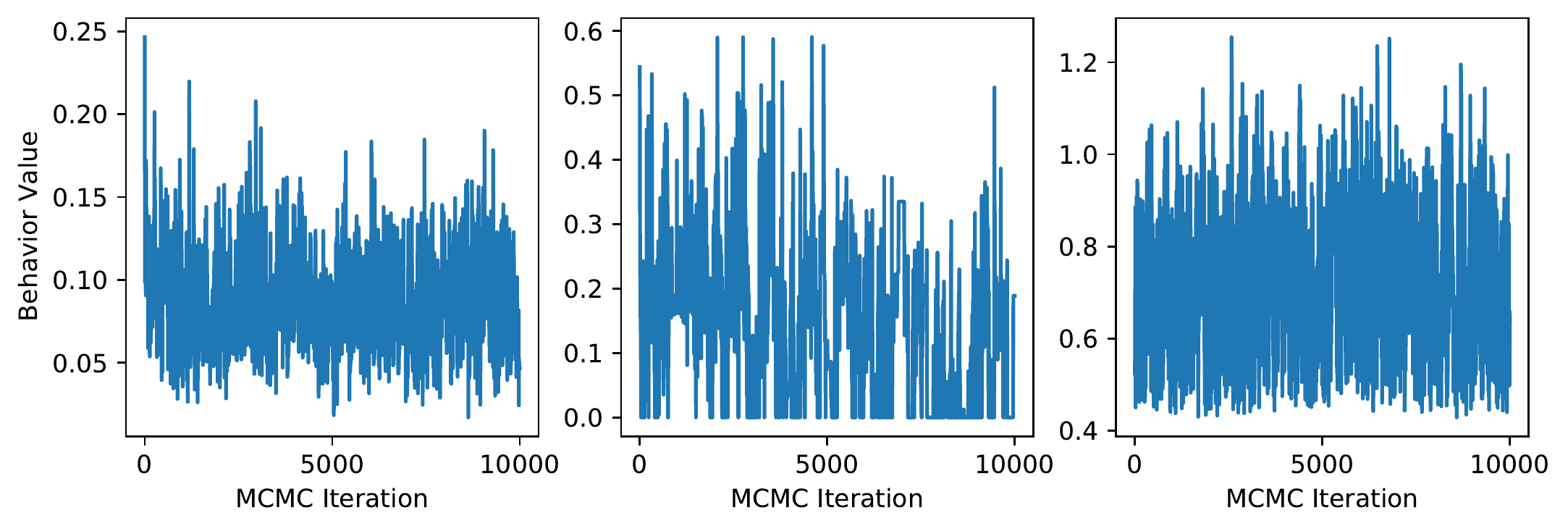}
    \vspace{-0.1in}
    \caption{The sampled behavior values for three MCMC chains. From left to right, the three panels show DS min straight-line deviation on 2D navigation, RRT min straight-line deviation on 2D navigation and RL min end-effector movement on 7DoF arm reaching. The visualization confirms that 10,000 iterations with 5,000 burn-ins are more than sufficient to find representative samples. }
    \label{fig:num-samples-app}
\end{figure}

Each sampling run collected 10,000 samples, with the first 5,000 discarded as burn-in. On a consumer-grade computer with a single GeForce GTX 1080 GPU card (for neural network-based controllers), the sampling generally takes around 1 to 3 hours. The number of samples and burn-ins are selected fairly conservatively to ensure representativeness, as Fig.~\ref{fig:num-samples-app} plots the sampled behavior values in the chain for three analyses and confirms that these numbers are more than sufficient to ensure proper mixing. Note that \modelname{} is designed to be an offline analysis tool as opposed to be used for real-time sample generation, and therefore several hours of runtime would be acceptable in most cases. Furthermore, MCMC sampling is embarrassingly parallel by simply using multiple chains concurrently, with the only overhead cost being the discarded burn-in samples.

\section{2D Environment Details}
\label{app:rbf-env-vis}

\begin{figure}[!htb]
    \centering
    \includegraphics[width=\textwidth]{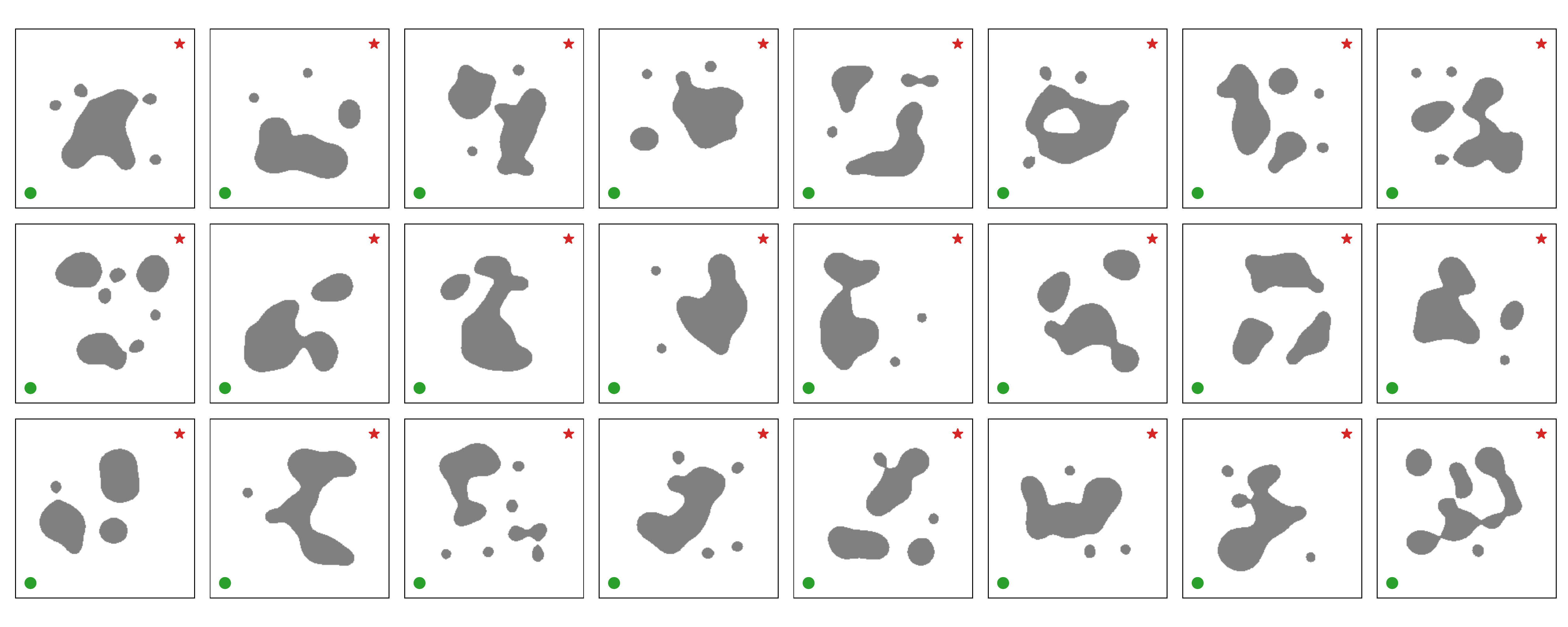}
    \caption{An assortment of randomly generated RBF 2D environments, providing a sense of the diversity generated with this formulation. The green dots are the environment starting points and the red stars are navigation targets. We show DS modulation for the first three environments in Fig.~\ref{fig:modulation-2D-supp}.}
    \label{fig:environment-variety-supp}
\end{figure}

In this domain, the environment is the area defined as $[x_\text{min}, x_\text{max}] \times [y_\text{min}, y_\text{max}]$. The goal is to navigate from $[x_\text{start}, y_\text{start}]$ to $[x_\text{goal}, y_\text{goal}]$. We define a flexible environment representation as a summation of radial basis function (RBF) kernels centered at so-called \emph{obstacle points.} Specifically, given $N_O$ obstacle points $\vec p_1, \vec p_2, ..., \vec p_{N_O}\in \mathbb R^2$, the environment is defined as 
\begin{align}
    e(\vec p) = \sum_{i=1}^{N_O} \exp \left(-\gamma ||\vec p-\vec p_i||_2^2\right), 
\end{align}
and each point $\vec p$ is an obstacle if $e(\vec p) > \eta$, for $\eta < 1$ to ensure each obstacle point $\vec p_i$ is exposed as an obstacle. 
Our environments are bounded by $[{-1.2}, 1.2] \times [{-1.2}, 1.2]$, and the goal is to navigate from  $[{-1}, {-1}]$ to $[1, 1]$. $N_O=15$ and $p_i$ coordinates are sampled uniformly in $x_i, y_i \in [{-0.7}, 0.7]$. A smaller $\gamma$ and $\eta$ makes the obstacles larger and more likely to be connected; we choose $\gamma=25$ and $\eta=0.9$. Fig.~\ref{fig:environment-variety-supp} shows random obstacle configurations demonstrating high diversity in this environment. We also implement a simple simulator: given the current robot position $[x, y]$ and the action $[\Delta x, \Delta y]$, the simulator clamps $\Delta x, \Delta y$ to the range of [-0.03, 0.03], and then moves the robot to $[x + \Delta x, y + \Delta y]$ if there is no collision, and otherwise simulates a frictionless inelastic collision (i.e. compliant sliding) that moves the robot tangent to the obstacle. Fig.~\ref{fig:environment-variety-supp} depicts a randomly selected assortment of 2D environments. These environments demonstrate the flexibility and diversity of the RBF environment definition. 

\section{Implementation Details of 2D Navigation Controllers}
\label{app:2dnav-controller}

\subsection{IL Controller}

The imitation learning controller is a memoryless policy implemented as a fully connected neural network with two hidden layers of 200 neurons each and ReLU activations. The input is 18 dimensional, with two dimensions for the current $(x, y)$ position of the robot, and 16 dimensions for a lidar sensor in 16 equally-spaced directions, with a maximum range of 1. The network predicts the heading angle $\theta$, and the controller operates on the action of $[\Delta x, \Delta y]=[0.03\cos\theta, 0.03\sin\theta]$. 

The network is trained on smoothed RRT trajectories. Specifically, we use the RRT controller to find and discretize a trajectory. Then the smoothing procedure repeatedly replaces each point by the mid-point of its two neighbors, absent collisions. When this process converges, each point on the trajectory becomes one training data point. 

Since only local observations are available and the policy is memoryless, the robot may get stuck in obstacles, which happens in approximately $10\%$ of the runs. In addition, while the output target is continuous, a regression formulation with mean-squared error (MSE) loss is inappropriate, due to multimodality of the output. For example, when the robot is facing an obstacle, moving to either left or right would avoid it, but if both directions appear in the dataset, the MSE loss would drive the prediction to be the average, resulting in a head-on collision. This problem has been recognized in other robotic scenarios such as grasping \cite{zhou20176dof} and autonomous driving \cite{xu2017end}. We follow the latter to treat this problem as classification with 100 bins in the $[0, 2\pi]$ range. 

\subsection{DS Controller}
\label{app:2d-ds}

For the DS controller, there are two technical challenges in using the modulation \citep{huber2019avoidance} on our RBF-defined environment. First, we need to identify and isolate each individual obstacle, and second, we need to define a $\Gamma$-function for each obstacle. 

To find all obstacles, we discretize the environment into an occupancy grid of resolution $150\times 150$ covering the area of $[-1.2, 1.2] \times [-1.2, 1.2]$. Then we find connected components using flood fill, and each connected component is taken to be an obstacle. 

\begin{figure}[!htb]
    \centering
    \vspace{0.1in}
    \includegraphics[width=0.2\columnwidth,trim={120px 80px 160px 80px},clip]{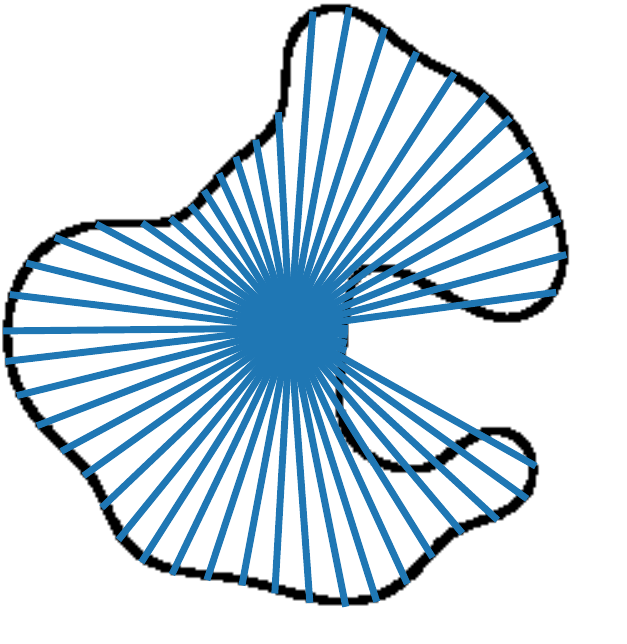}
    \includegraphics[width=0.2\columnwidth,trim={120px 80px 160px 80px},clip]{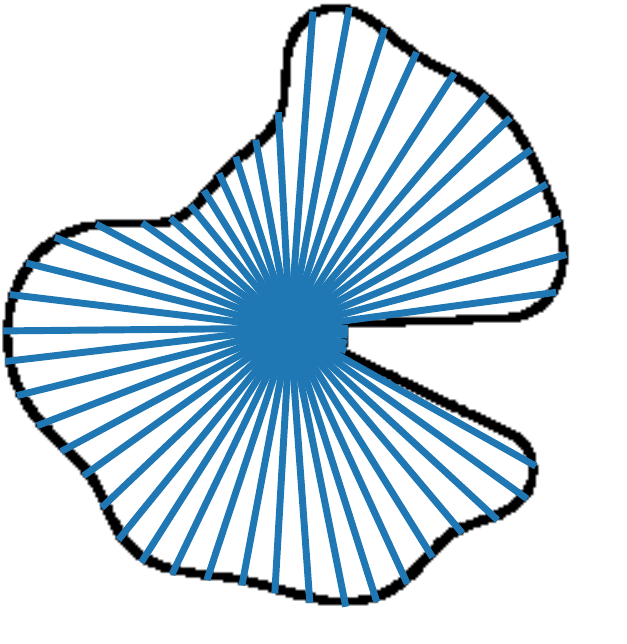}
    \caption{Left: an obstacle which is not star-shaped. Some radial lines extending from the obstacle's reference point cross the boundary of the obstacle twice. Right: the same obstacle, modified to instead be star-shaped.}
    \label{fig:star-shaped-explainer-supp}
\end{figure}

To define a $\Gamma$-function for each obstacle, we first choose the reference point as the center of mass of the connected component. Then we cast 50 rays in 50 equally spaced directions from the reference point and find the intersection point of each ray with the boundary of the connected component. Finally, we  connected those intersections in sequence and get a polygon. In case of multiple intersection points, we take the farthest point as vertex of the polygon, essentially completing the non-star-shaped obstacle to be star-shaped, as shown in Fig. \ref{fig:star-shaped-explainer-supp}. 

Given an arbitrary point $\vec x$, we define 
\begin{align}
    \Gamma(\vec x) = \frac{||\vec x - \vec r||}{||\vec i - \vec r||}, 
\end{align}
where $\vec r$ is the reference point and $\vec i$ is the intersection point with the polygon of the ray from $\vec r$ in $\vec x - \vec r$ direction. It is easy to see that this $\Gamma$ definition satisfies all three requirements for $\Gamma$-functions listed in App. \ref{app:ds-general}. 

\begin{figure}[!htb]
    \centering
    \includegraphics[width=0.2\textwidth,trim={80px 20px 80px 2px},clip]{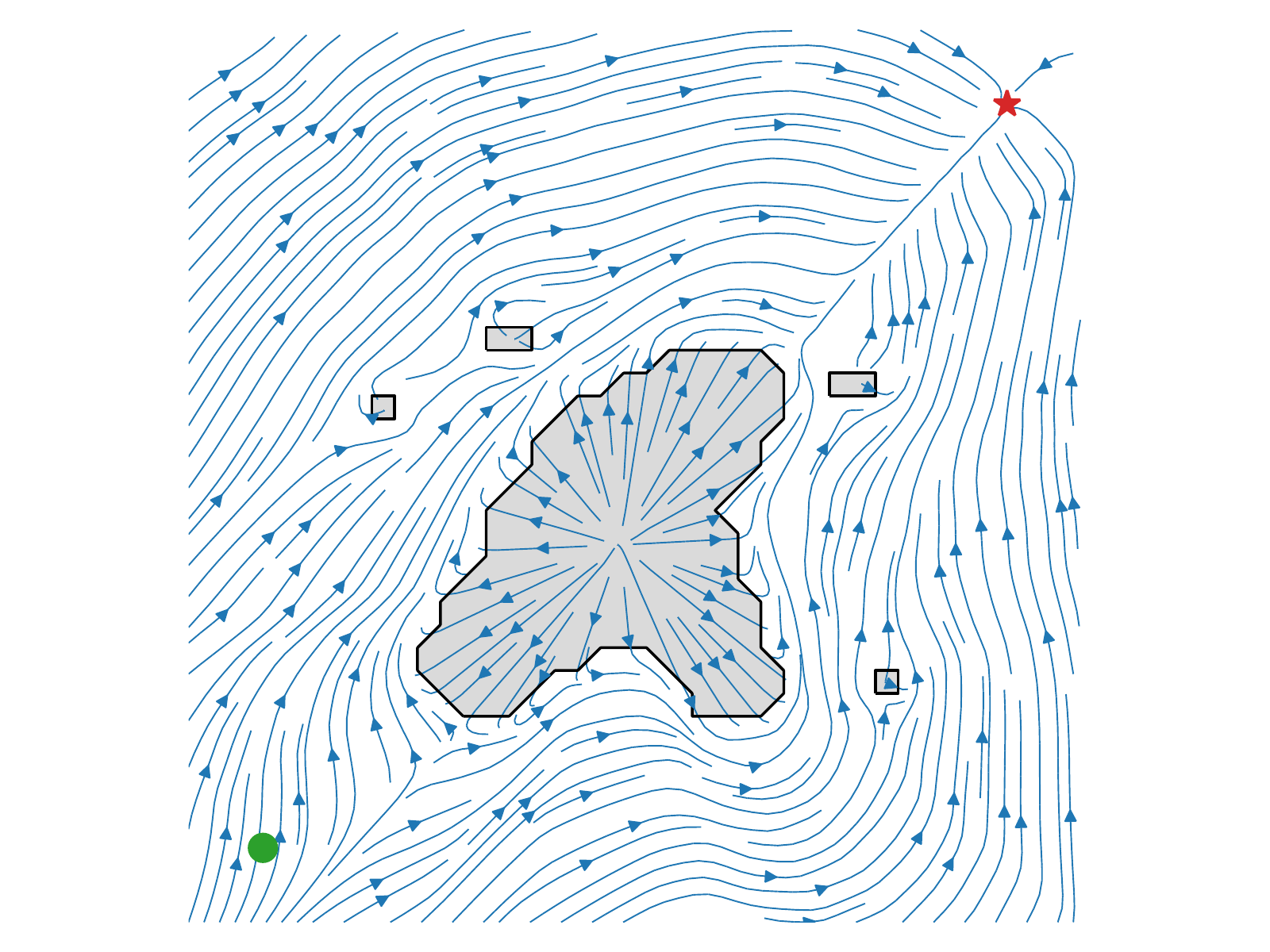}
    \hspace{1mm}
    \includegraphics[width=0.2\textwidth,trim={80px 20px 80px 20px},clip]{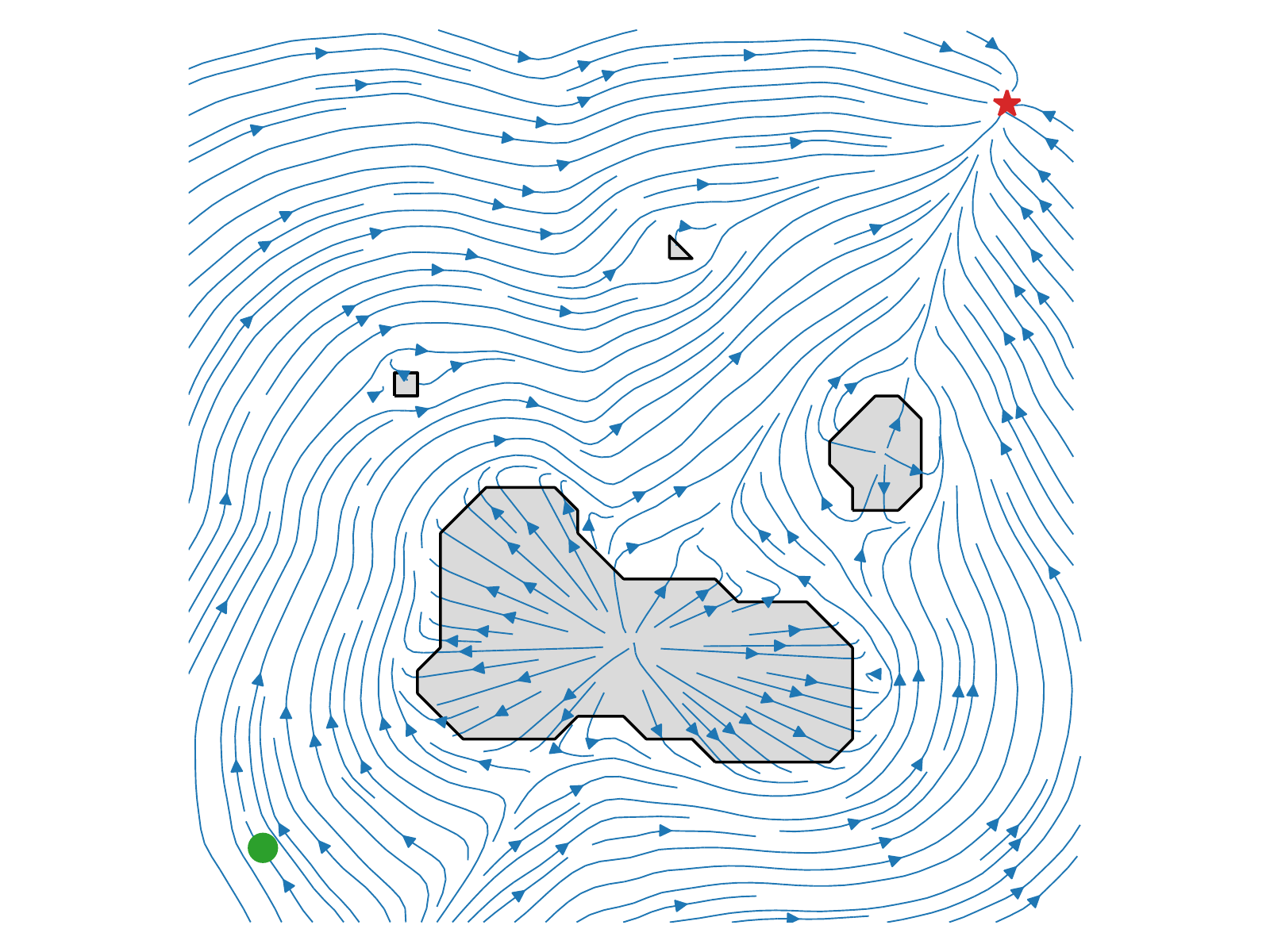}
    \hspace{1mm}
    \includegraphics[width=0.2\textwidth,trim={80px 20px 80px 20px},clip]{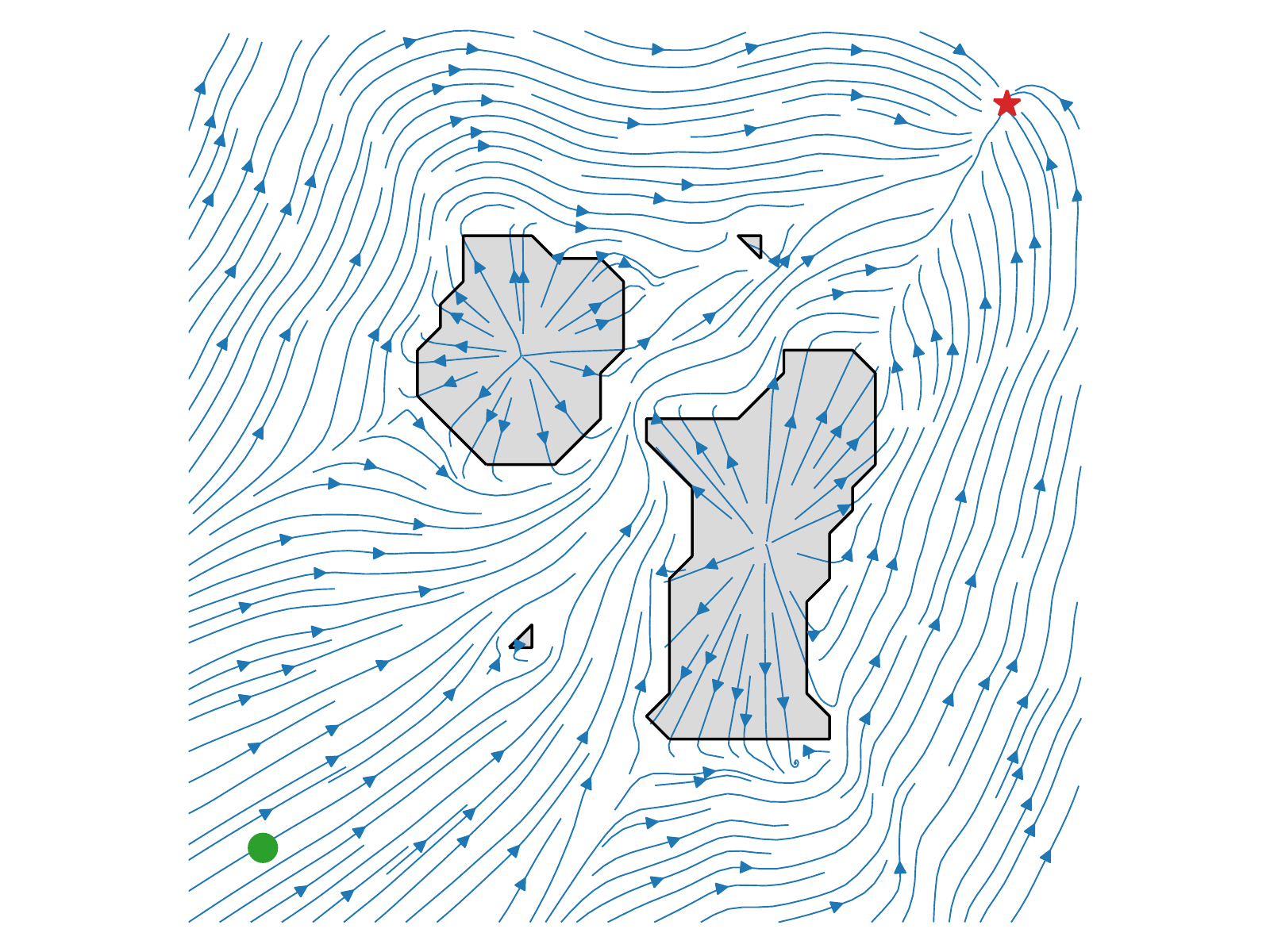}
    \caption{Streamlines showing the modulation effect of the dynamical system for three 2D navigation tasks. The environments correspond to the first three examples of Fig.~\ref{fig:environment-variety-supp}. Green dots are starting positions and red stars are navigation targets.}
    \vspace{0.1in}
    \label{fig:modulation-2D-supp}
\end{figure}

Finally, to compensate for numerical errors in the process (e.g. approximating obstacles with polygons), we define the control inside obstacle to be the outward direction, which helps preventing the robot from getting stuck at obstacle boundaries in practice. Three examples of DS modulation of the 2D navigation environment are shown in Fig.~\ref{fig:modulation-2D-supp}.

\section{Additional Results for 2D Navigation}
\label{app:2dnav-exp}

\begin{figure}[!htb]
    \centering
    \includegraphics[width=0.7\columnwidth,trim={10px 5px 5px 0px},clip]{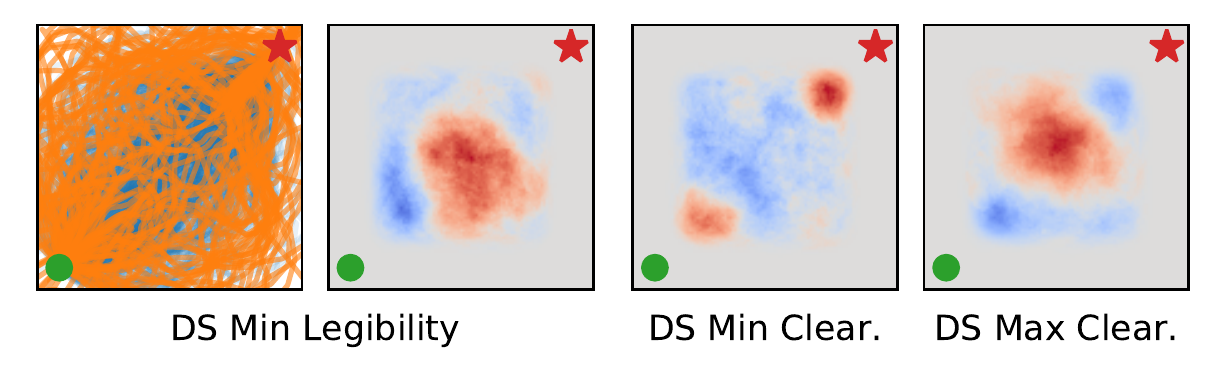}
    \caption{Left: trajectories and obstacle configurations from sampling minimal DS legibility. Right: obstacle configurations for minimizing and maximizing DS obstacle clearance. These examples show how obstacle positions affect the legibility and clearance behaviors.}
    \vspace{0.1in}
    \label{fig:2d-misc-behavior}
\end{figure}

\paragraph{Legibility} We define the instantaneous legibility as the cosine similarity between the current robot direction and the direction to target $\vec x^*$, $V(\vec x)=\dot{\vec x} \cdot (\vec x^*-\vec x) / (||\dot{\vec x}||\cdot||\vec x^*-\vec x||)$, with the intuition that a particular run may be confusing to users if the robot does not often align to the target. Though this quantity is bounded by $[-1, 1]$, a general legibility definition may not be. Thus, we use the maximal mode of \modelname{} to find DS trajectories and obstacle configurations that achieve \textit{minimal} legibility, by negating $V(\vec x)$ first. 
The left two panels of Fig.~\ref{fig:2d-misc-behavior} present the samples. As expected, most trajectories take large detours due to the presence of obstacles in the center. 

\paragraph{Obstacle Clearance} We take $V(\vec x) = \min_{\vec x_o\in \mathcal O}||\vec x - \vec x_o||$. For the DS, we sample two posteriors to maximize and minimize this behavior. As shown in the right two panels of Fig.~\ref{fig:2d-misc-behavior}, when minimizing obstacle clearance, we see clusters of obstacles in close proximity to the starting and target positions, such that the robot is forced to navigate around them. When maximizing obstacle clearance, we instead see central clusters of obstacles, such that the robot can avoid them by bearing hard left or right.

\section{Implementation Details of 7DoF Arm Reaching Controllers}
\label{app:7dof-controller}

\subsection{RRT Controller}
\label{app:3d-rrt}

Since the target location is specified in the task space, we first find the target joint space configuration using inverse kinematics (IK). The initial configuration starts with the arm positioned down on the same side as the target. If the IK solution is in collision, we simulate the arm moving to it using position control, and redefine the final configuration at equilibrium as the target (i.e. its best effort reaching configuration). We solve the IK using Klamp't \cite{hauser2016robust}. 

\subsection{RL Controller}
\label{app:3d-rl}

The RL controller implements the proximal policy gradient (PPO) algorithm \cite{schulman2017proximal}. The state space is 22-dimensional and consists of the following: 
\begin{itemize}[leftmargin=0.2in, topsep=0pt, itemsep=0pt, parsep=0pt]
    \item 7D joint configuration of the robot, 
    \item 3D position of the end-effector, 
    \item 3D roll-pitch-yaw of the end effector, 
    \item 3D velocity of the end-effector, 
    \item 3D position of the target, 
    \item 3D relative position from the end-effector to the target. 
\end{itemize}
The action is 7-dimensional for movement in each joint, which is capped at $[-0.05, 0.05]$. 

Both the actor and the critic are implemented with fully connected networks with two hidden layers of 200 neurons each, and ReLU activations. The action is parametrized as Gaussian where the actor network predicts the mean, and 7 standalone parameters learns the log variance for each of the 7 action dimensions. At test time, the policy deterministically outputs the mean action given a state.

\subsection{DS Controller}
\label{app:3d-ds}

For the DS controller in 7DoF arm reaching, we face the same challenges as in 2D navigation: defining an appropriate $\Gamma$-function for the obstacle configuration that holds the three properties introduced by \citet{huber2019avoidance} (listed in App. \ref{app:ds-general}). Additionally, the DS modulation technique does not consider the robot's morphology, end-effector shape, or workspace limits because it only modulates the state of a point-mass. Thus, we implement several adaptations. First, we modulate the 3D position of the tip of the end-effector. The desired velocity of the end-effector tip, given by the modulated linear controller, is then tracked by the 7DoF arm via the same position-level IK solver as the RRT controller.

\begin{figure}[!htb]
    \centering
    \includegraphics[width=0.49\columnwidth]{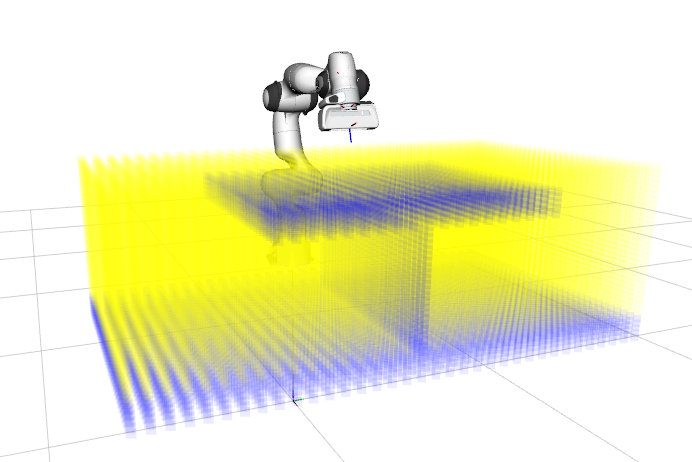}
    \includegraphics[width=0.49\columnwidth,trim={30px 50px 30px 50px},clip]{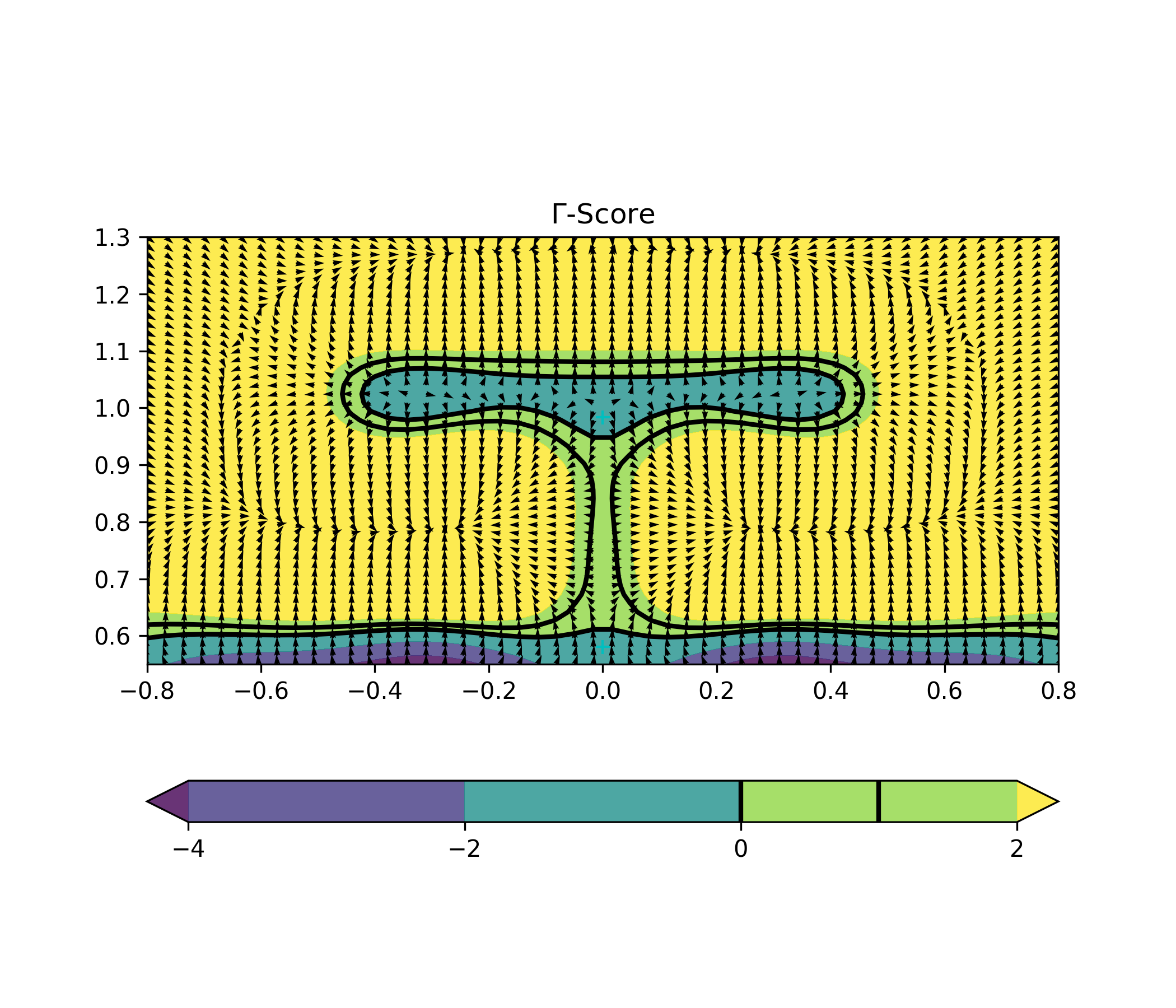}
    \caption{Left: the division of 3D space as either containing an obstacle or free space. This data is used to train an SVM, which acts as an interpolator. The classification scores of the SVM are used as the $\Gamma$ function for this 3D reaching task. Right: a 2D slice showing the smoothed $\Gamma$ scores.}
    \label{fig:ds-3d-gamma}
\end{figure}

Second, we used a support vector machine (SVM) to learn the obstacle boundary from a list of points in the obstacle and free spaces, an approach originally proposed by \citet{mirrazavi2018unified}. Then the decision function of the SVM is used as the $\Gamma$-function. As shown in Fig. \ref{fig:ds-3d-gamma}, we discretize the 3D workspace of the robot and generate a dataset of points in the obstacle space as negative class and those in the free space as positive class.

\begin{figure}[!htb]
    \centering
    \includegraphics[width=0.49\columnwidth,trim={30px 30px 30px 30px},clip]{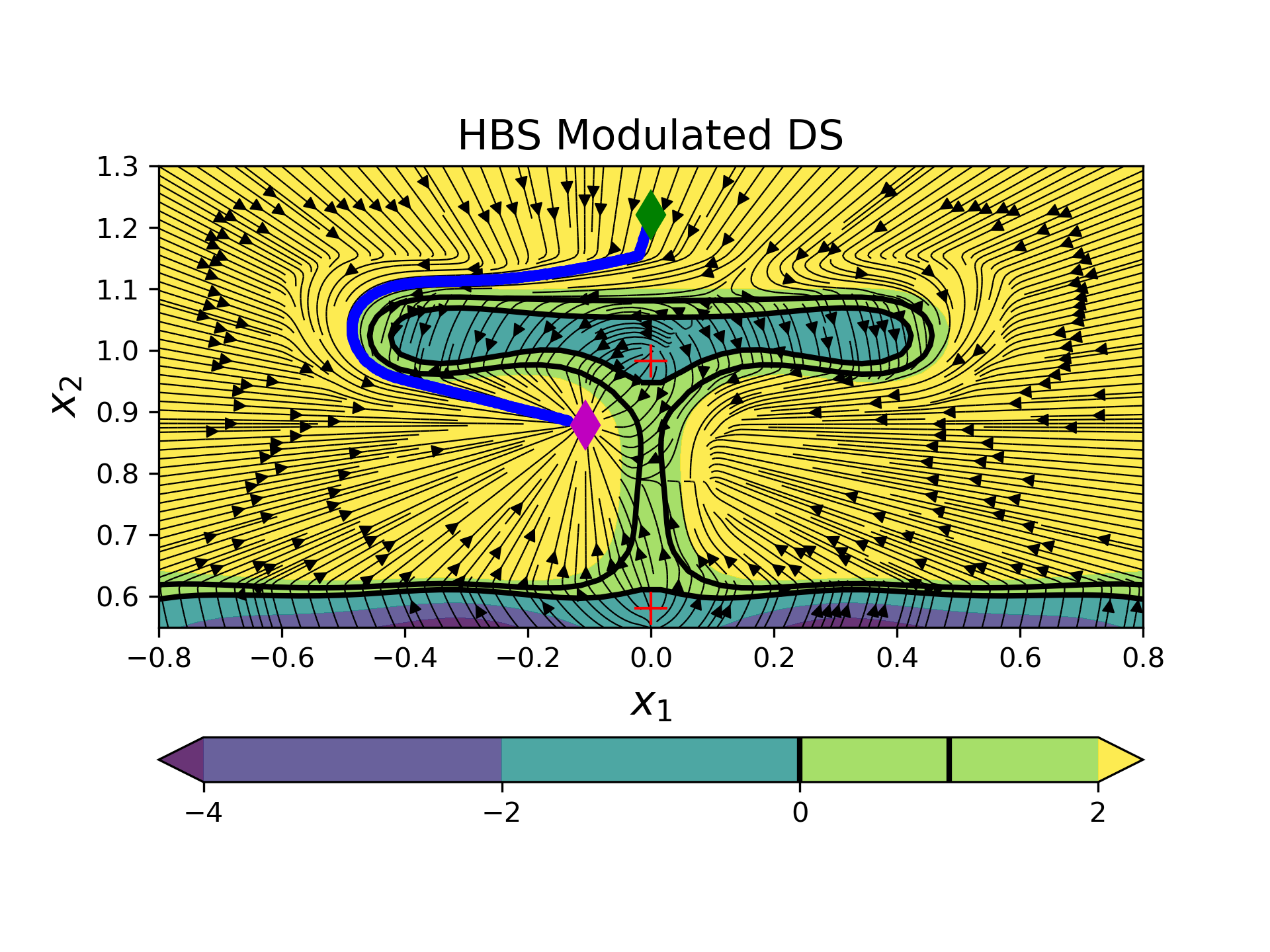}
    \includegraphics[width=0.49\columnwidth,trim={30px 30px 30px 30px},clip]{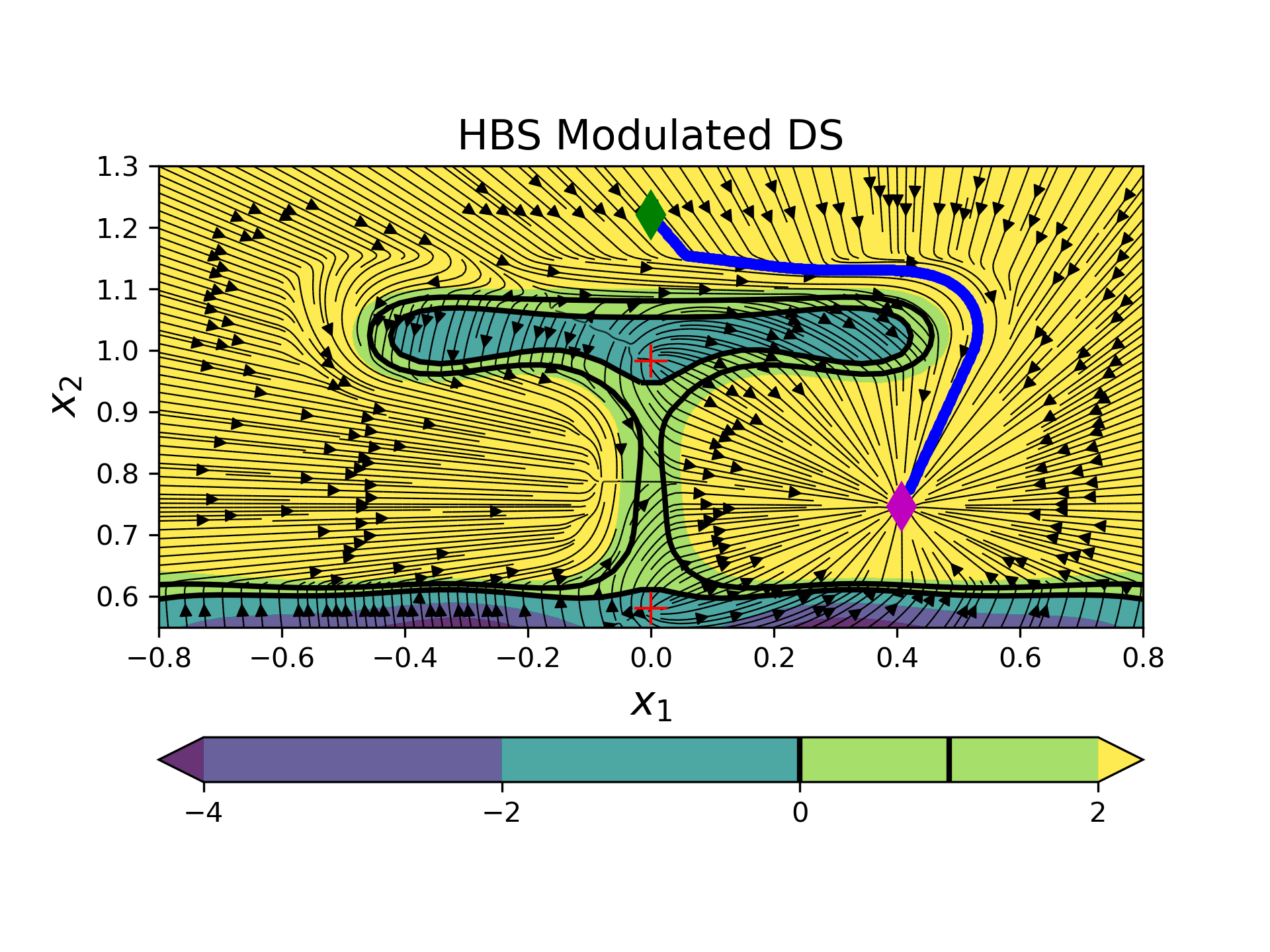}
    \caption{Cross-sections showing streamlines of the dynamical system modulation effect for two distinct targets in the 3D reaching task. Red crosses indicate reference points. Green diamond is the initial position of the end-effector for all experiments.}
    \label{fig:ds-3d-modulations}
\end{figure}

Using the radial basis function (RBF) kernel $K(\vec x_1,\vec x_2) = e^{-\gamma||\vec x_1 - \vec x_2||^2}$, with kernel width $\gamma$, the SVM decision function $\Gamma(\vec x)$ has the following form:
\begin{equation}
\begin{aligned}
\label{eq:Gamma}
\Gamma(\vec x) = \sum_{i=1}^{N_{sv}}\alpha_i y_i K(\vec x,\vec x_i)+b =\sum_{i=1}^{N_{sv}}\alpha_i y_i e^{-\gamma||\vec x - \vec x_i||^2}+b, 
\end{aligned}
\end{equation}
and the equation for $\nabla\Gamma (\vec x)$ is naturally derived as follows:
\begin{equation}
\begin{aligned}
\label{eq:dGamma}
\nabla\Gamma(\vec x) = \sum_{i=1}^{N_{sv}}\alpha_i y_i \dfrac{\partial K(\vec x,\vec x_i)}{\partial \vec x}
=-\gamma\sum_{i=1}^{N_{sv}}\alpha_i y_i e^{-\gamma||\vec x - \vec x_i||^2}(\vec x - \vec x_i) .
\end{aligned}
\end{equation}
In Eq.~\ref{eq:Gamma} and \ref{eq:dGamma}, $\vec x_i\ (i = 1, ..., N_{sv})$ are the support vectors from the training dataset, $y_i$ are corresponding collision labels ($-1$ if position is collided, $+1$ otherwise), $0 \leq \alpha_i \leq C$ are the weights for support vectors and $b\in \mathbb{R}$ is decision rule bias. Parameter $C\in \mathbb{R}$ is a penalty factor used to trade-off between errors minimization and margin maximization. We empirically set the hyper-parameters of the SVM to $C=20$ and $\gamma=20$. Parameters $\alpha_i$ and $b$ and the support vectors $\vec x_i$ are estimated by solving the optimization problem for the soft-margin kernel SVM using \texttt{scikit-learn}. Using this learned $\Gamma$-function, Fig.~\ref{fig:ds-3d-modulations} shows two examples of the modulated trajectory.

Finally, given a desired modulated 3D velocity for the end-effector tip, $\dot{\vec x}_M = \vec u_M(\vec x)$, we compute the next desired 3D position by numerical integration:
\begin{equation}
    \vec x_{t+1} = \vec x_{t} + \vec u_M(\vec x_{t})\Delta t 
\end{equation}
where $\vec x_{t},\vec x_{t+1} \in \mathbb{R}^3$ are the current and next desired 3D position of the tip of the end-effector and $\Delta t = 0.03$ is the control loop time step. $\vec x_{t+1}$ is then the target in Cartesian world space coordinates that defines the objective of the position-based IK solver implemented in Klamp't \cite{hauser2016robust}.

\section{Additional Results for 7DoF Arm Reaching}
\label{app:ds-improvement}
\label{app:3d-additional-vis}

\paragraph{Details on the DS Improvement}
The DS controller provides guarantees of convergence to a target in the space where modulation is applied (i.e. task-space in our experiments). To adopt this controller for obstacle avoidance with a robot manipulator, \citet{huber2019avoidance} simplifies the robot to a spherical shape with center at the end-effector of a 7DOF arm. This translates to considering the robot as a zero-mass point in 3D space but with the boundaries of the obstacles (described by $\Gamma$-functions) expanded by a margin with the size of the radius of the sphere. 

Since the shape of the Franka robotic hand is rectangular (6.3 $\times$ 20.7 $\times$ 14cm) fitting a sphere with the radius of the longest axis will over-constrain the controller and drastically reduce the target regions inside the table dividers. We thus implemented the obstacle clearances by extruding the edges of the top table divider by half of the length of the robot's end-effector (10cm) and the width of the divider by half of the height (7cm). Intuitively, this should be enough clearance to avoid the robot's end-effector colliding with the table dividers. However, when coupling the DS controller with the IK solver to control the 7DoF arm, we noticed that the success rate was below $15\%$, whereas the success rate is $100\%$ when controlling the end-effector only. We then sampled, via \modelname{}, the target locations for the minimal final end-effector distance to target and noticed that all of the successful runs were located on the left-side of the partition (Fig.~\ref{fig:3d-min-ee-and-ds-improvement} center right). 

Since the DS controller approach does not consider collision avoidance in joint-space, in a constrained environment, the robot's forearm or elbow might get stuck on the edges of the table divider---even though the end-effector is avoiding collision. Due to the asymmetric kinematic structure of the robot arm, it is more prone to these situations on the right side of the table divider. Such an insight is not easy to discover as one must understand how the robot will behave in joint space based on its kinematic structure and the low-level controller used (position-based IK). We thus extended the edge extrusions to 20cm. This change improved the controller success rate and behavior drastically as shown in (Fig.~\ref{fig:3d-min-ee-and-ds-improvement} rightmost).

\paragraph{Legibility} We define legibility of reaching to the target on one side of the vertical divider as the average negative distance that the end effector moves in the other direction, $V(\vec x)=-\max(\tilde {\vec x}_1, 0)$, where $\tilde {\vec x}_1=\vec x_1$ if target is on the left, or $\tilde {\vec x}_1=-\vec x_1$ otherwise, and $\vec x_1$ is the $x$-coordinate of the robot end effector with right in the positive direction. We find target locations that are minimally legible and apply the maximal inference mode on the maximum distance measure. 

\begin{figure}[!htb]
    \centering
    \includegraphics[width=0.5\columnwidth, trim=2.4in 2.2in 0 0, clip]{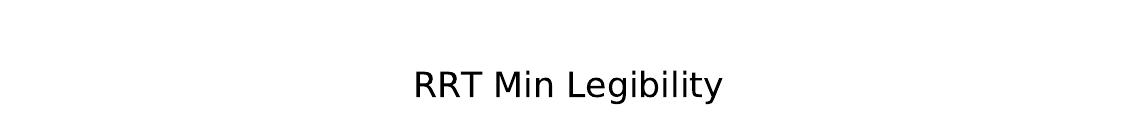}
    \includegraphics[width=0.5\columnwidth, trim=2.4in 0.5in 0 0.8in, clip]{figures/3d_rrt_max_illegibility.pdf}
    \caption{Posterior samples showing minimal legibility behavior for RRT.}
    \vspace{0.1in}
    \label{fig:3d-rrt-illegibility}
\end{figure}

We did not find any illegible motions from RL controllers for 2,000 targets, which is mostly expected since the RL reward is distance to the target. For RRT, however, since we do not use an optimal formulation \cite[e.g.][]{karaman2011sampling, hauser2016asymptotically} or perform post-hoc smoothing, the controller is expected to frequently exhibit low legibility. Fig.~\ref{fig:3d-rrt-illegibility} plots the posterior target locations and trajectories. The target locations leading to illegible motions are spread out mostly uniformly on the right, but concentrated in far-back area on the left, consistent with our findings on the asymmetry of configuration space. The trajectory plot confirms the illegibility. 

\section{Future Work}
\label{app:future-work}
There are multiple directions to extend and complement \modelname{} for better usability and more comprehensive functionality. First, while we only used \modelname{} on individual controllers, future work can readily extend it to \textit{compare two controllers} by defining behavior functions that take in the task and two trajectories, one from each controller, and compute differential statistics. For example, this could be used to find road conditions that lead to increased swerving behavior of a new AV controller, compared to the existing one. Such testing is important to gain a better understanding of \textit{model updates}~\citep{bansal2019updates}, and is particularly necessary for ensuring that these updates do not unintentionally introduce new problems. 

In addition, sometimes it is important to understand particular trajectories sampled by \modelname{}. For example, which sensor input (e.g. lidar or camera) is most important to the current action (e.g. swerving)? Why does the controller take one action rather than another (e.g. swerving rather than braking)? Preliminary investigation into this explainable artificial intelligence (XAI) problem in the context of temporally extended decision making has been undertaken~\citep{greydanus2018visualizing, zahavy2016graying}, but various issues with existing approaches have been raised~~\citep{atrey2019exploratory, zhou2021feature} and future research is needed to address them. 

Finally, an important step before actual deployment is to design appropriate user interfaces to facilitate the two-way communication between \modelname{} and end-users. In one direction, the user needs to specify the behavior of interest, and it would be desirable for it to involve as little programming as possible, especially for non-technical stakeholders. In the other direction, \modelname{} needs to present the sample visualization, and potentially model explanations as described above, for users to inspect. Here, it is important for the information to be accurate but at the same time not overwhelming.

\end{document}